\RequirePackage{fix-cm}
\documentclass[twocolumn]{svjour3}          
\smartqed  
\usepackage{graphicx}
\usepackage{gensymb}
\usepackage{amsmath}
\usepackage{amsfonts}
\usepackage{multirow}
\usepackage{cite}
\usepackage{xcolor}

\usepackage{hyperref}
\hypersetup{colorlinks=true,citecolor=[rgb]{0.00 0.0 1.00},linkcolor=[rgb]{0.00 0.0 1.00},urlcolor=[rgb]{0.00 0.0 1.00},hypertexnames=false}

\newcommand{\Fig}{Fig.}
\newcommand{\fig}{Fig.}

\iffalse
\newcommand{\SpecOne}{1}
\newcommand{\SpecTwo}{2}
\newcommand{\SpecThree}{3}
\newcommand{\SpecFour}{4}
\newcommand{\SpecFive}{5}
\newcommand{\SpecSix}{6}

\else
\newcommand{\SpecOne}{6}
\newcommand{\SpecTwo}{1}
\newcommand{\SpecThree}{2}
\newcommand{\SpecFour}{3}
\newcommand{\SpecFive}{4}
\newcommand{\SpecSix}{5}

\fi

\newcommand{\SuppTab}[1]{Supp. Table~\mbox{#1}}

\begin{document}

\title{Automatic Annotation of Hip Anatomy in Fluoroscopy for Robust and Efficient 2D/3D Registration}

\titlerunning{Accepted for IPCAI 2020}        

\author{Robert B. Grupp\textsuperscript{1}         \and
   		    Mathias Unberath\textsuperscript{1} \and
		    Cong Gao\textsuperscript{1} \and \\
            Rachel A. Hegeman\textsuperscript{2} \and 
            Ryan J. Murphy\textsuperscript{3} \and
            Clayton P. Alexander\textsuperscript{4} \and  
            Yoshito Otake\textsuperscript{5} \and
            Benjamin A. McArthur\textsuperscript{6,7} \and
            Mehran Armand\textsuperscript{2,4,8} \and
            Russell H. Taylor\textsuperscript{1}
            }

\authorrunning{Accepted for IPCAI 2020} 

\institute{R. B. Grupp\\
			  \email{grupp@jhu.edu}\\
			  \textsuperscript{1}Department of Computer Science, Johns Hopkins University, Baltimore, MD, USA\\
			  \textsuperscript{2}Research and Exploratory Development Department, Johns Hopkins University Applied Physics Laboratory, Laurel, MD, USA\\
			  \textsuperscript{3}Auris Health, Inc., Redwood City, CA, USA\\
			  \textsuperscript{4}Department of Orthopaedic Surgery, Johns Hopkins Medicine, Baltimore, MD, USA\\
			  \textsuperscript{5}Graduate school of Information Science, Nara Institute of Science and Technology, Ikoma, Nara, Japan\\
			  \textsuperscript{6}Department of Surgery and Perioperative Care, Dell Medical School, University of Texas, Austin, TX, USA\\
			  \textsuperscript{7}Texas Orthopedics, Austin, TX, USA\\
			  \textsuperscript{8}Department of Mechanical Engineering, Johns Hopkins University, Baltimore, MD, USA
}

\date{Received: date / Accepted: date}
%

\maketitle

\begin{abstract}
\setlength{\itemindent}{0.5em}
\item[\textbf{Purpose}]
Fluoroscopy is the standard imaging modality used to guide hip surgery and is therefore a natural sensor for computer-assisted navigation.
In order to efficiently solve the complex registration problems presented during navigation, human-assisted annotations of the intraoperative image are typically required.
This manual initialization interferes with the surgical workflow and diminishes any advantages gained from navigation.
In this paper we propose a method for fully automatic registration using anatomical annotations produced by a neural network.
\item[\textbf{Methods}]
Neural networks are trained to simultaneously segment anatomy and identify landmarks in fluoroscopy.
Training data is obtained using a computationally-intensive, intraoperatively incompatible, 2D/3D registration of the pelvis and each femur.
Ground truth 2D segmentation labels and anatomical landmark locations are established using projected 3D annotations.
Intraoperative registration couples a traditional intensity-based strategy with annotations inferred by the network and requires no human assistance.
\item[\textbf{Results}]
Ground truth segmentation labels and anatomical landmarks were obtained in 366 fluoroscopic images across 6 cadaveric specimens.
In a leave-one-subject-out experiment, networks trained on this data obtained mean dice coefficients for left and right hemipelves, left and right femurs of 0.86, 0.87, 0.90, and 0.84, respectively.
The mean 2D landmark localization error was 5.0 mm.
The pelvis was registered within 1\degree~for 86\% of the images when using the proposed intraoperative approach with an average runtime of 7 seconds.
In comparison, an intensity-only approach without manual initialization, registered the pelvis to 1\degree~in 18\% of images.
\item[\textbf{Conclusions}]
We have created the first accurately annotated, non-synthetic, dataset of hip fluoroscopy.
By using these annotations as training data for neural networks, state-of-the-art performance in fluoroscopic segmentation and landmark localization was achieved.
Integrating these annotations allows for a robust, fully automatic, and efficient intraoperative registration during fluoroscopic navigation of the hip.
\keywords{Landmark Detection \and Semantic Segmentation \and 2D/3D Registration \and X-ray Navigation \and Orthopaedics}
\end{abstract}

\section{Introduction}\label{sec:intro}
Minimally invasive surgical interventions of the hip manipulate, modify, or augment anatomical structures which are hidden or not reliably visible~\cite{woerner2016visual}.
Clinicians commonly use intraoperative fluoroscopy in order to overcome this occlusion and ascertain the poses of anatomy, surgical instruments, or artificial implants.
However, mental interpretation of these images is a difficult task and subject to an extensive learning curve~\cite{slotkin2015accuracy,troelsen2009surgical}.
Computer-assisted navigation systems ease this burden by tracking relevant objects and reporting their poses within the context of a surgical plan or scenario.
Systems leveraging fluoroscopy have been developed for total hip arthroplasty~\cite{kelley2009role}, hip resurfacing~\cite{belei2007fluoroscopic}, cement injection~\cite{malan2016fluoroscopy}, and osteotomies of the acetabulum~\cite{grupp2019pose} or proximal femur~\cite{gottschling2005intraoperative}.

In order to report object poses accurately, fluoroscopic navigation systems rely on 2D/3D registrations between intraoperative 2D images and the appropriate 3D models~\cite{markelj2012review}.
Large errors in pose estimates may occur when a registration is not initialized sufficiently close to the actual pose of an object.
Quality initializations are derived from some manual human input, often through annotated landmark locations in the fluoroscopic image.
Although these systems report favorable navigation-related results, the manual initialization of processing may interrupt surgical workflows and negatively affect patient outcomes due to increased operating time or blood loss.

Convolutional neural networks (CNNs) have excelled at detecting landmarks and performing semantic segmentation when sufficiently large annotated datasets are available for training~\cite{newell2016stacked,shelhamer2017fully}.
However, since existing large-scale hip datasets have focused on 3D image modalities and pre and postoperative radiography, rather than intraoperative fluoroscopy~\cite{otake2018construction}, applications of CNNs to fluoroscopy have been mostly limited to recognizing surgical instruments and tools~\cite{miao2016cnn,ambrosini2017fully,breininger2018intraoperative,gao2019localizing}.

Several authors have coupled image segmentation with landmark estimation using multi-task networks and achieved favorable results.
By reusing segmentation features from an encoder-decoder style network for the computation of landmark heatmaps, Laina was able to automatically annotate segmentation labels and landmark locations of tools used in laparoscopy and retinal microsurgery~\cite{laina2017concurrent}.
Gao also leveraged this approach for the localization of a dextrous continuum manipulator in fluoroscopy~\cite{gao2019localizing}.
Kordon demonstrated that a CNN, trained from 149 manually annotated preoperative radiographs, could successfully segment the four bones of the knee joint, and locate two anatomical landmarks and a surgically relevant line~\cite{kordon2019multi}.

Using a large collection of simulated fluoroscopy, Bier trained CNNs to annotate anatomical landmarks of the pelvis~\cite{bier2019learning}.
When evaluated on five sequences of actual fluoroscopy across two cadaveric specimens, mean annotation errors of 12-24 mm in the detector plane were reported.
Pelvis poses were estimated using these annotations, yielding reprojection errors of 14-34 mm for other landmarks not learned by the network.
Their work was extended in~\cite{esteban2019towards}, whereby each network was fine-tuned on simulated fluoroscopy for a specific patient of interest.
The approach was evaluated by estimating landmark locations in previously unseen simulated images, and using these estimates to produce quality initializations for 2D/3D registration.
No analysis on actual fluoroscopy was conducted in~\cite{esteban2019towards}.

In this paper, we propose a method for 2D/3D registration of hip anatomy that simultaneously combines image intensities with higher-level landmark and segmentation features, making it robust against large initial offsets from actual object poses.
Segmentation labels and landmark annotations are produced by a CNN similar in architecture to those found in~\cite{gao2019localizing} and~\cite{laina2017concurrent}.
Contrary to~\cite{bier2019learning,esteban2019towards,gao2019localizing}, we train our networks using smaller datasets of \textit{actual} fluoroscopy and achieve state-of-the-art results on clinically relevant data.
Annotated fluoroscopy for \textit{training} is semi-automatically obtained using a computationally expensive 2D/3D registration, with runtimes on the order of several minutes per image.

The novel contributions of this paper are:
\begin{itemize}
  \item A semi-automatic, \textit{offline}, pipeline for creating the first annotated training dataset of semantically segmented individual bone structures and anatomical landmark locations in actual hip fluoroscopy,
  \item A demonstration that CNN models, trained using small datasets of less than 400 annotated images, can achieve state-of-the-art-results for the tasks of semantic segmentation and landmark localization in actual hip fluoroscopy,
  \item An \textit{online}, \textit{intraoperative}, registration strategy, leveraging image intensities and CNN-features, that is fully automatic, requires no initialization from a user, and completes in an order of seconds.
\end{itemize}
\begin{figure*}
\begin{centering}
\includegraphics[width=0.75\textwidth]{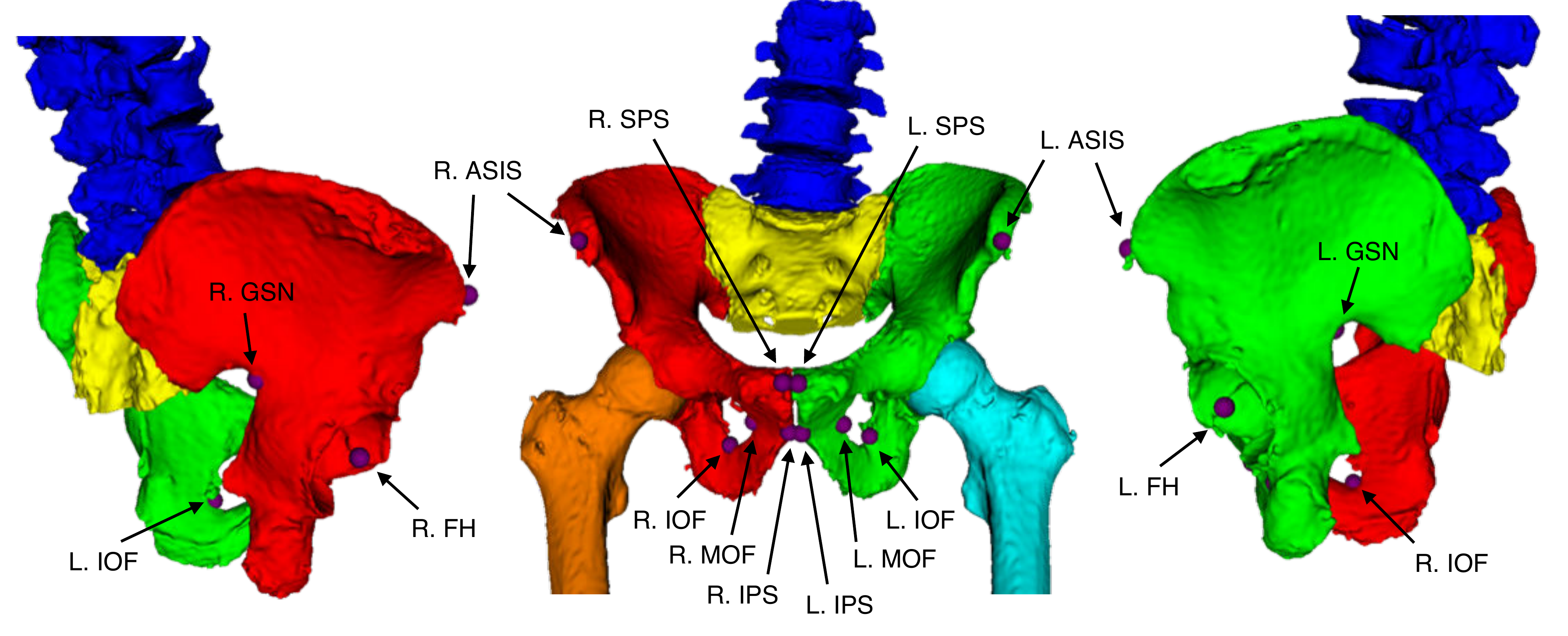}
\caption{Three views of the 6 anatomical structures and 14 landmarks to be annotated in 2D fluoroscopy.
	    		 All landmarks are bilateral with left (L.) and right (R.) denoted.
			 The L. hemipelvis is shown in green, the R. hemipelvis in red, L. femur in cyan, R. femur in orange, vertebrae in blue, and upper sacrum in yellow.
			 Each landmark is overlaid as a purple sphere.
			 }
\label{fig:lands_example}
\end{centering}
\end{figure*}
\section{Methods}\label{sec:methods}
We now describe the details of the data preprocessing, the methods for creating an annotated, \textit{training}, dataset of hip fluoroscopy, the CNN architecture, and the intraoperative registration strategy.
The reader is referred to Appendix~\ref{sec:supp_methods} for details regarding the lower-level parameters used for the registration pipelines and network training.
\subsection{Data Preprocessing}\label{sec:methods_data}
Using the procedure described in~\cite{grupp2019pose}, lower torso 3D CT scans are resampled to have 1 mm isotropic spacing.
Segmentations of the pelvis, femurs, and vertebrae are obtained semi-automatically.
A total of 14 landmarks are manually annotated in 3D:
the left and right (L./R.) centers of the femoral head (FH),
L./R. greater sciatic notches (GSN),
L./R. inferior obturator foramen (IOF),
L./R. medial obturator foramen (MOF),
L./R. superior pubis symphysis (SPS),
L./R. inferior pubis symphysis (IPS),
and the L./R. anterior superior iliac spine (ASIS).
These landmarks were previously identified as being useful for obtaining initial registration estimates of the pelvis~\cite{grupp2019pose}.
The anterior pelvic plane (APP) coordinate system for each specimen is defined using the L./R. ASIS and L./R. SPS landmarks~\cite{nikou2000description}, and is later used to estimate nominal anterior/posterior (AP) poses and as a reference coordinate frame during registration.
Segmentations of each hemipelvis and sacrum are separated from the full pelvis segmentation, and any sacrum labels inferior to the sacroiliac joint are discarded.
\Fig~\ref{fig:lands_example} shows an example 3D visualization of the individual bone surfaces and the anatomical landmarks.

Fluoroscopy is collected with a Siemens CIOS Fusion mobile C-arm with $30 \times 30\text{ cm}^2$ detector. Images are $1536 \times 1536$ pixels with $0.194$ mm pixel spacings.
Each image is cropped by 50 pixels along each border to remove collimator artifacts and intensity values are log-corrected (``bone is bright'').
\subsection{2D/3D Registration}\label{sec:methods_regi_2d3d}
Our approach to 2D/3D registration of single-view fluoroscopy and CT builds upon the multiple-resolution, multiple-component, 2D/3D, intensity-based registration pipeline introduced in~\cite{grupp2019pose}.
The registration problem of finding the rigid poses of the pelvis ($\theta_P$), left femur ($\theta_{LF}$), and right femur ($\theta_{RF}$) with respect to a single fluoroscopic view, $I$, is defined by the optimization problem \eqref{eq:regi_problem}, where $\mathcal{P}$ indicates a projection operator creating digitally reconstructed radiographs (DRRs), $\mathcal{S}$ indicates a similarity measure between DRRs and fluoroscopy, $\mathcal{R}$ is a regularization penalizing implausible poses, and $\lambda \in [0,1]$ is a tuning parameter.
\begin{equation} \label{eq:regi_problem}
\begin{gathered}
	\min_{\theta_{P}, \theta_{LF}, \theta_{RF} \in SE(3)} \lambda \mathcal{S} \left( \mathcal{P} \left(  \theta_{P}, \theta_{LF}, \theta_{RF} \right), I \right) + \\
		\hfill \left( 1 - \lambda \right) \mathcal{R} \left( \theta_{P}, \theta_{LF}, \theta_{RF} \right)
\end{gathered}
\end{equation}
%
In this paper, $\mathcal{S}$ is defined as the weighted sum of normalized cross-correlations of 2D image gradients computed over image patches~\cite{grupp2018patch}.
For all registrations using regularization, $\lambda = 0.9$.
\subsection{Training Dataset Creation}\label{sec:methods_dataset_creation}
The \textit{training} dataset of annotated fluoroscopy images is constructed using an automatic 2D/3D registration of the pelvis and both femurs.
Once anatomy is registered to each image, the 3D segmentation labels and landmarks are propagated to 2D.
Since this registration is performed ``offline,'' we use a computationally expensive combination of global search strategies, followed by several local strategies.
Manual inspection is performed so that images corresponding to failed registrations are pruned from the dataset.
It should be emphasized that, although this registration is automatic and global, the amount of computation precludes it from intraoperative application.

%
An attempt is first made to register the pelvis using a mixture of the Differential Evolution~\cite{storn1997differential}, exhaustive grid search, Particle Swarm~\cite{shi1998modified}, Covariance Matrix Adaptation: Evolutionary Search (CMA-ES)~\cite{hansen2001completely}, and Bounded Optimization by Quadratic Approximation (BOBYQA)~\cite{powell2009bobyqa} optimization strategies at multiple resolutions.
Using a combination of the CMA-ES and BOBYQA strategies, the left and right femurs are registered once the pelvis is registered.
The rotation components of the left femur and right femurs are independently estimated, keeping the pelvis fixed at its current pose estimate.
Next, simultaneous optimization over the rigid poses of the pelvis and both femurs is performed.
Multiple resolution levels are used throughout this process, with downsampling factors along each 2D dimension ranging from $32\times$ to $4 \times$.
For each of the preceding registrations uniform patch weightings were applied for $\mathcal{S}$.

The 2D location of each landmark is obtained by projecting the corresponding 3D landmark onto the detector.
When a landmark projects outside the detector region, it is identified as not visible in the image.

Each 2D pixel is labeled as the anatomy for which the corresponding source-to-detector ray intersects.
Discrete labels are used to assign a single class of anatomy to each pixel.
Femurs are given highest precedence in labeling: any ray/femur intersection yields a label of the corresponding femur.
Hemipelves have the next highest precedence, with any rays intersecting both hemipelves assigned a label corresponding to the hemipelvis closer to the X-ray source.
Vertebrae intersections are given next precedence, followed by the upper sacrum. All remaining pixels are assigned to background.

The 2D labels and landmarks for each projection are manually inspected and verified.
\begin{figure*}
\begin{centering}
\includegraphics[width=0.725\textwidth]{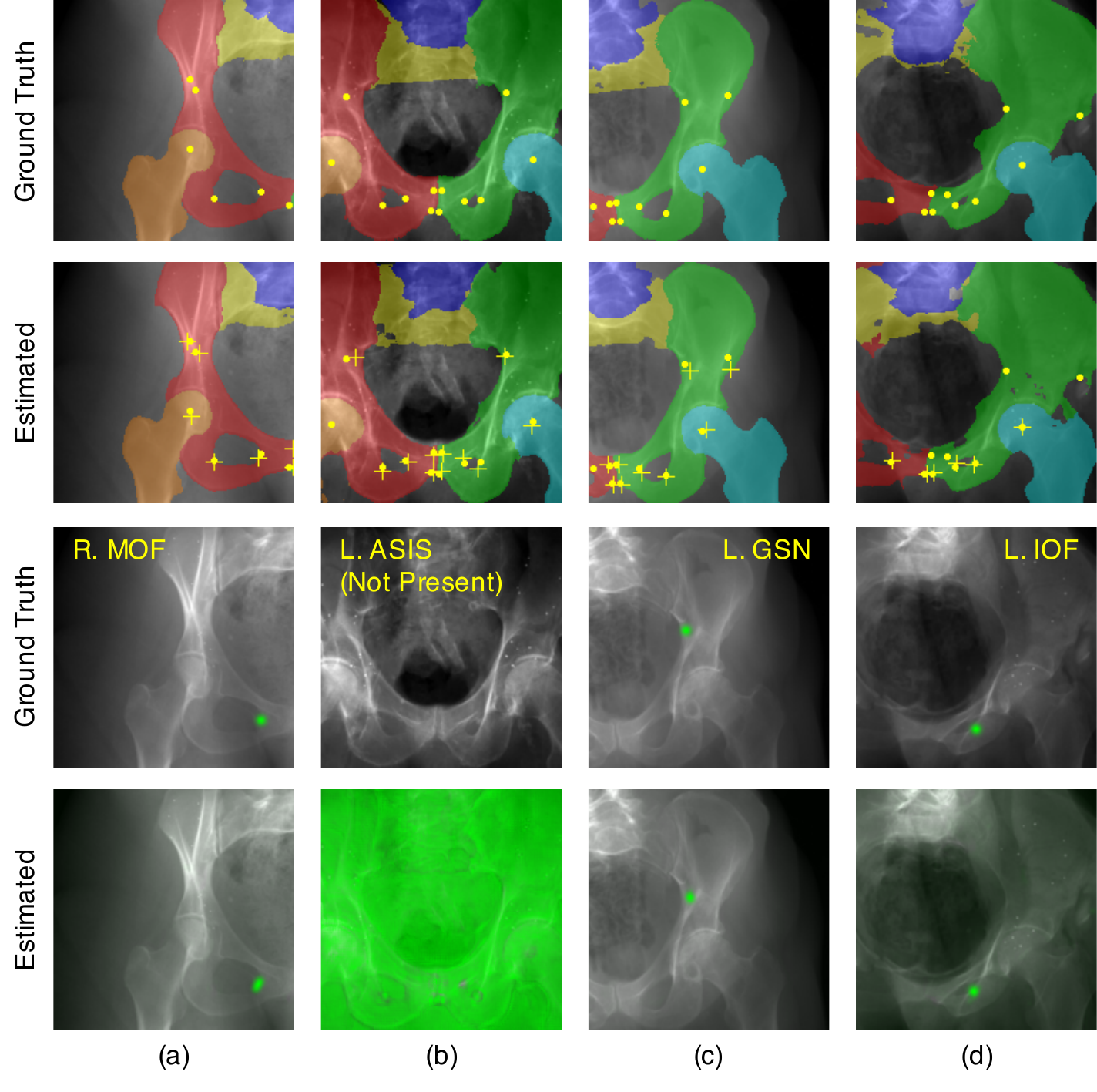}
\caption{Example annotations of four specimens. 
			 The top row shows the ground truth segmentation labels for each object overlaid onto the fluoroscopic images, along with the landmark locations as yellow circles.
			 The colors of each object correspond to those from Fig. \ref{fig:lands_example}.
			 CNN estimates are shown in the second row, with ground truth landmark locations shown as yellow circles and estimated locations shown as yellow crosshairs (+).
			 Missed detections are indicated by a circle without a corresponding cross.
			 Ground truth heatmaps for the R. MOF, L. ASIS, L. GSN, and L. IOF, in (a), (b), (c), and (d), respectively, are overlaid and shown in the third row.
			 Estimated heatmaps for these landmarks are shown in the bottom row.
			 The heatmap shown in (b) highlights a successful no detection report for L. ASIS.
			 }
\label{fig:gt_and_est_segs_lands_example}
\end{centering}
\end{figure*}
\subsection{Network Architecture and Training}\label{sec:methods_network}
In constructing our network, we follow the approach described by \cite{gao2019localizing} and \cite{laina2017concurrent}, appending segmentation and landmark heatmap network paths after an encoder-decoder structure.
Supp. Figs.~\ref{fig:unet_block}, \ref{fig:unet_enc_dec}, and~\ref{fig:network} describe the network architecture used in this work.
For the encoder-decoder in this paper, we adopt a 6 level U-Net \cite{ronneberger2015u} design with 32 features at the top level and 1024 features at the bottom.
Our implementation is fully-convolutional with learned 2x2 convolutions of stride 2 for downsampling, and transposed convolutions for upsampling.

The segmentation path follows directly from the original U-Net design. 
The differentiable dice score \cite{milletari2016v} is computed for each class and then averaged.
This value is bounded, taking on values in $[0 , 1]$, with larger values indicating a higher quality segmentation.

Segmentation features prior to soft-max are concatenated with the features output from the encoder-decoder, and passed through two 1x1 convolutions to obtain a feature map where each channel estimates the heatmap of a landmark.

Ground truth heatmaps for each landmark location are defined by a symmetric 2D normal distribution with mean value equal to the landmark location and standard deviations of $\sigma = 3.88$ mm in each direction.
The value of $\sigma$ was subjectively chosen to approximate the variance found in manual landmark annotation.
Each heatmap is set to be identically zero when the landmark is not visible.
Examples of ground truth heatmaps are shown in the third row of \Fig~\ref{fig:gt_and_est_segs_lands_example}.
Heatmap loss is computed using the average normalized-cross correlation (NCC) between each ground truth heatmap and the corresponding estimate.
This term is bounded, taking on values in $[-1,1]$, with larger positive values indicating stronger correlation between ground truth heatmaps and estimated heatmaps.

By scaling and shifting the average NCC value into the range of $[0,1]$, the heatmap loss may be weighted equally to the dice term without any additional hyper-parameter tuning.
Finally, the combined dice and heatmap terms are negated (for minimization).

Networks are trained using stochastic gradient descent, with an initial learning rate of 0.1, Nesterov momentum of 0.9, weight decay of 0.0001, and a batch size of five images.
Training and validation data sets are obtained by applying a random 85\%/15\% split to the data not used for testing.
Test data sets are comprised of images collected from a single specimen, and no images derived from this specimen are present in the training and validation data.
Extensive online data augmentation is applied to each image with probability 0.5.
If an image is to be augmented, the intensities are randomly inverted, random noise is added to the image, the contrast is randomly adjusted, a random 2D affine warp is applied, and a random number of regions are corrupted with very large amounts of noise.
Each image is normalized to have zero mean and standard deviation one before input into the network.
Training is run for a maximum of 500 epochs and the learning rate is multiplied by $0.1$ after validation loss plateaus.
The network expects images of size $192\times192$ pixels, and fluoroscopy data is downsampled $8\times$ in each dimension, accordingly.
PyTorch 1.2 was used to implement, train, and test the networks.
\subsection{Extracting Landmark Locations} \label{sec:methods_extract_lands}
Both, segmentations and heatmaps, are used to estimate anatomical landmark locations.
Candidate locations of the FH landmarks are restricted to pixels labeled as the corresponding femur, and all remaining landmarks are restricted to locations labeled as the corresponding hemipelvis.
Restricting candidate locations in this way avoids possible false alarms when the ipsilateral landmark is not in the view and a large heatmap intensity is located about the contralateral landmark.
The final proposed location of each landmark is defined as the candidate location with maximal heatmap intensity.
In order to distinguish between the cases of landmark detection, no detection, and spurious heatmap values, a $25^2$ pixel region of the estimated heatmap, centered around the proposed location, is matched against the 2D symmetric normal distribution template of a detection at the center of the region.
A detection is reported when NCC between the two regions is greater than 0.9, and no detection is reported otherwise.
\subsection{Intraoperative Registration}\label{sec:methods_intraop_regi}
The intraoperative registration strategies in this paper attempt to solve \eqref{eq:regi_problem} in a similar fashion as the method used for construction of the training data set: the pelvis is registered first, followed by optimizations of each femur's rotation, followed by a simultaneous optimization over the rigid poses of all objects.

\textit{Method 1}: A naive approach for efficient automatic registration uses only intensity information, with uniform patch weightings and no regularization applied.
The single landmark initialization described in~\cite{grupp2019pose} is used to calculate an initial AP pose of the pelvis, aligning the 3D centroid of the L. ASIS, R. ASIS., L. SPS, and R. SPS with the center of the image.

\textit{Method 2}: However, a great deal of information about the 2D image is known, courtesy of the segmentation and landmark localizations produced by the CNN.
A less naive approach uses detected landmarks to solve the PnP problem~\cite{hartley2003multiple} and automatically initializes an intensity-based registration.
The segmentation is used to apply non-uniform patch weightings in $\mathcal{S}$, and soft-bounds are applied through a regularization on pose parameters.

\textit{Method 3}: Instead of treating landmark features and intensity features separately, the detected landmark locations may be incorporated into a robust reprojection regularizer for intensity-based registration.
The regularizer is defined in \eqref{eq:intraop_landmark_regularization}, with the $l^\text{th}$ landmark location in 3D is denoted by $p^{(l)}_\text{3D}$, and corresponding estimated 2D location, $p^{(l)}_\text{2D}$.
\begin{equation} \label{eq:intraop_landmark_regularization}
		\mathcal{R} \left( \theta_{P} \right) = \frac{1}{2\sigma_\ell^2} \sum_{l = 1}^{N_L} \left\| \mathcal{P} \left( p^{(l)}_\text{3D} ; \theta_{P} \right) - p^{(l)}_\text{2D} \right\|_2^2
\end{equation}
As with the PnP approach, non-uniform patch weightings are applied using the segmentation.
Using one of the estimated 2D landmark locations, the single landmark initialization is used to calculate an initial AP pose of the pelvis.

Pelvis registrations first use a CMA-ES optimization, followed by the BOBYQA strategy at a finer resolution without patch weightings or regularization.
When using patch weightings, patches centered at pixels labeled as either hemipelvis are given uniform weight and the remaining patches are weighted zero.
Next, femur registration proceeds identically to that used during construction of the training data set, except for the case of patch weighting.
When registering the individual femurs and using patch weightings, patches centered at pixels labeled as either hemipelvis or either femur are given uniform weight and all remaining patches are weighted zero.
Multiple resolution levels are used, with either $8\times$ or $4 \times$ downsampling applied.
\begin{figure}
\begin{centering}
\includegraphics[width=0.9\columnwidth]{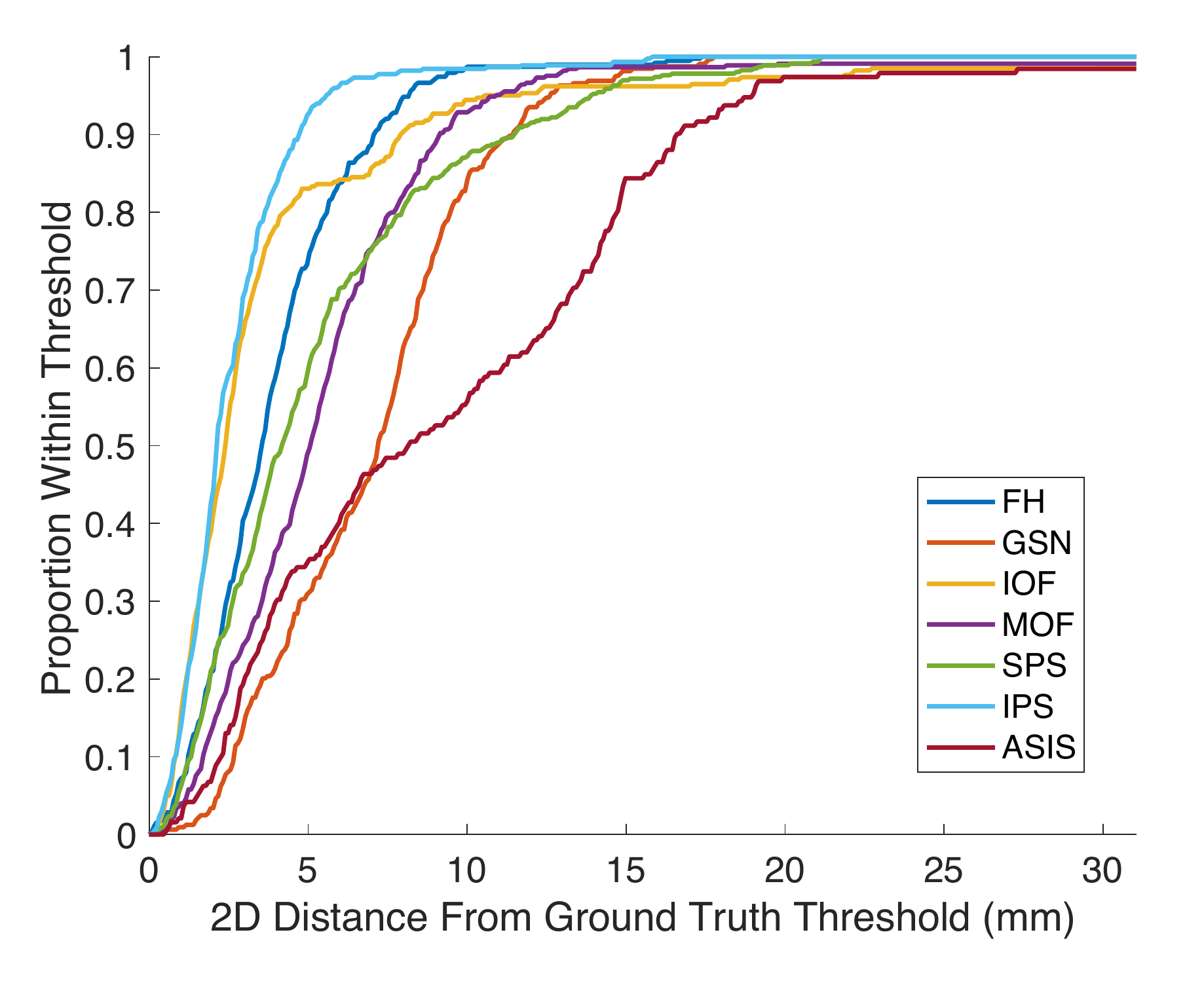}
\caption{A plot of 2D landmark detection accuracy given various thresholds in mm.
			 The bilateral cases for each landmark are combined in this plot.
			 }
\label{fig:lands_acc_plot}
\end{centering}
\end{figure}
\section{Experiments and Results}\label{sec:results}
\subsection{Data Collection and Training Dataset Creation}\label{sec:results_dataset_creation}
Lower torso CT scans were obtained for three male and three female cadaveric specimens with median age 88 and ranging 57-94 years.
Each CT scan was semi-automatically segmented and 3D landmarks were manually digitized.
A total of 399 fluoroscopic images were collected at various C-arm poses.
The ``offline'' \textit{training} data set registration pipeline was run on each image and a total of 366 images were verified to have registered successfully.
For each successfully registered image, manual annotations were made for each femur indicating whether enough of the bone was visible to use for future registration evaluation.
These counts are also broken down across each specimen in \SuppTab{\ref{tab:training_data}}.

Across all specimens, the intraoperative femur poses differed from their poses during preoperative CT scanning by an average of $15.3\degree \pm 7.4\degree$ and $0.7 \pm 0.6$ mm.
For specimens \SpecFour~and \SpecOne, there were no images which had sufficient views of the left femur needed to evaluate registration.
For specimen \SpecFour, only two images had sufficient views of the right femur needed to evaluate registration, and the femur was in the same pose for these two views.
In the images which had sufficient views for femoral registration, more than two poses were observed for each femur of specimens \SpecTwo, \SpecThree, \SpecFive, and \SpecSix.

Poses recovered from successful registrations in this phase were treated as ground truth during intraoperative registrations.
Examples of generated 2D ground truth annotations are shown in the top row of \Fig~\ref{fig:gt_and_est_segs_lands_example}.
The mean total registration time per image was 4 minutes using a NVIDIA Tesla P100 (PCIe) GPU.
\subsection{Segmentation and Landmark Localization}\label{sec:results_seg_lands}
A total of six networks were trained in a leave-one-specimen-out experiment.
For each network, the training and validation data consisted of all labeled images from five specimens and all labeled images from the remaining specimen were used as test data.
Across all test images, mean dice coefficients of $0.86 \pm 0.20$, $0.87 \pm 0.18$, $0.90 \pm 0.24$, $0.84 \pm 0.31$, $0.74 \pm 0.19$, and $0.63 \pm 0.13$ were obtained for the left hemipelvis, right hemipelvis, left femur, right femur, vertebra, and upper sacrum, respectively.
A listing of dice coefficients for each object of each specimen is shown in \SuppTab{\ref{tab:results_dice_scores}}.
The average landmark 2D localization error was $5.0 \pm 5.2$ mm in the detector plane.
Table~\ref{tab:results_landmark_errors} lists the average landmark errors, false negative rates, and false positive rates for each landmark.
\Fig~\ref{fig:lands_acc_plot} shows a plot of localization error thresholds and corresponding correct detection rates.
%
The mean time for segmentation and landmark detection per image was $24.0 \pm 0.4$ milliseconds using a NVIDIA Tesla P100 (PCIe) GPU.
\begin{table*}
\centering
\caption{Landmark detection errors across all trained networks for each landmark.
			 }
\label{tab:results_landmark_errors}       
\begin{tabular}{l r r r r}
\hline\noalign{\smallskip}
\multirow{2}{*}{Landmark} & \multicolumn{2}{c}{Error} & \multirow{2}{*}{False Negative Rate} & \multirow{2}{*}{False Positive Rate} \\ \cline{2-3}
                                          & \multicolumn{1}{c}{Pixels} & \multicolumn{1}{c}{mm}                    &  &  \\
\noalign{\smallskip}\hline\noalign{\smallskip}
L. FH        & $1.9 \pm 0.9$  & $3.0 \pm 1.5$  & 0.02  & 0.02     \\
R. FH       & $3.2 \pm 1.9$  & $5.0 \pm 3.0$  & 0.04  & 0.01     \\
L. GSN    & $4.4 \pm 2.0$  & $6.8 \pm 3.1$  & 0.20  & 0.00     \\
R. GSN    & $4.7 \pm 2.3$  & $7.3 \pm 3.6$  & 0.14  & 0.01     \\
L. IOF     & $2.8 \pm 4.0$  & $4.3 \pm 6.2$  & 0.23  & 0.01     \\
R. IOF     & $2.3 \pm 3.3$  & $3.5 \pm 5.1$  & 0.16  & 0.02     \\
L. MOF    & $3.7 \pm 3.2$  & $5.8 \pm 5.0$  & 0.17  & 0.04     \\
R. MOF    & $3.4 \pm 1.9$  & $5.2 \pm 3.0$  & 0.17  & 0.02     \\
L. SPS    & $3.1 \pm 2.4$  & $4.7 \pm 3.7$  & 0.27  & 0.02     \\
R. SPS    & $3.7 \pm 2.9$  & $5.8 \pm 4.5$  & 0.22  & 0.01     \\
L. IPS    & $1.9 \pm 1.6$  & $3.0 \pm 2.4$  & 0.17  & 0.02     \\
R. IPS    & $1.5 \pm 1.0$  & $2.3 \pm 1.6$  & 0.15  & 0.01     \\
L. ASIS    & $9.0 \pm 9.6$  & $14.0 \pm 14.9$  & 0.29  & 0.01     \\
R. ASIS    & $3.9 \pm 3.7$  & $6.0 \pm 5.7$  & 0.14  & 0.01     \\
All          & $3.2 \pm 3.4$  & $5.0 \pm 5.2$  & 0.17  & 0.01     \\
\noalign{\smallskip}\hline
\end{tabular}
\end{table*}
\iffalse
\begin{table*}
\caption{Pelvis and femur registration errors from successful pelvis registrations using the three intraoperative approaches.
			 Femur registrations errors are reported for all successful pelvis registrations which have sufficient visibility of a femur.}
\label{tab:results_regi_errors_all_specs}       
\begin{tabular}{l r r r r r r}
\hline\noalign{\smallskip}
 \multirow{2}{25pt}{Regi. Method} & \multicolumn{3}{c}{Pelvis Errors} & \multicolumn{3}{c}{Femur Errors} \\ \cline{2-7}
& \# Success & Rot. ($\degree$) & Trans. (mm) & \# & Rot. ($\degree$) & Trans. (mm) \\
\noalign{\smallskip}\hline\noalign{\smallskip}
1: Naive
  & 66 (18\%) &	 $0.1 \pm 0.1$ & $0.4 \pm 0.9$ & 22 & $0.4 \pm 0.3$ & $0.4 \pm 0.3$ \\
2: PnP Init.
  & 299 (82\%) &	 $0.1 \pm 0.2$ & $1.0 \pm 1.5$ & 183 & $1.4 \pm 3.4$ & $0.5 \pm 0.5$ \\
3: Combined
  & 313 (86\%) &	 $0.2 \pm 0.2$ & $1.4 \pm 2.0$ & 192 & $1.5 \pm 3.3$ & $0.6 \pm 0.7$ \\
\noalign{\smallskip}\hline
\end{tabular}
\end{table*}
\else
\begin{table*}
\centering
\caption{Pelvis and femur registration errors from successful pelvis registrations using the three intraoperative approaches and broken down by cadaver specimen.
			 Femur registrations errors are reported for all successful pelvis registrations which have sufficient visibility of a femur.}
\label{tab:results_regi_errors_all_specs}       
\begin{tabular}{l r r r r r r r}
\hline\noalign{\smallskip}
 \multirow{2}{22pt}{Regi. Method} & \multirow{2}{*}{Spec.} & \multicolumn{3}{c}{Pelvis Errors} & \multicolumn{3}{c}{Femur Errors} \\ \cline{3-8}
& & \# Success & Rot. ($\degree$) & Trans. (mm) & \# & Rot. ($\degree$) & Trans. (mm) \\
\noalign{\smallskip}\hline\noalign{\smallskip}
\multirow{7}{*}{\rotatebox{90}{1: Naive}}
& \SpecTwo    & 32 (29\%) &    $0.1 \pm 0.1$ & $0.3 \pm 0.2$ & 13   & $0.4 \pm 0.2$ & $0.3 \pm 0.3$ \\
& \SpecThree    & 15 (14\%) &	 $0.1 \pm 0.2$ & $0.8 \pm 1.9$ & 5   & $0.7 \pm 0.4$ & $0.4 \pm 0.5$ \\
& \SpecFour    & 1 (4\%)   &	 $ < 0.1 $ & $0.2 $ &  0    & \multicolumn{1}{c}{---} & \multicolumn{1}{c}{---} \\
& \SpecFive    & 4 (8\%)   &    $0.1 \pm 0.1$ & $0.4 \pm 0.4$ & 0   & \multicolumn{1}{c}{---} & \multicolumn{1}{c}{---} \\
& \SpecSix    & 13 (24\%)   &	 $0.1 \pm 0.1$ & $0.4 \pm 0.3$ & 3   & $0.4 \pm 0.2$ & $0.7 \pm 0.4$ \\
& \SpecOne    & 1 (4\%)   &	 $0.1$ & $0.3$ & 1     & $0.6$ & $0.2$ \\
& All  & 66 (18\%) &	 $0.1 \pm 0.1$ & $0.4 \pm 0.9$ & 22 & $0.4 \pm 0.3$ & $0.4 \pm 0.3$ \\
\noalign{\smallskip}\hline\noalign{\smallskip}
\multirow{7}{*}{\rotatebox{90}{2: PnP Init.}}
& \SpecTwo    & 99 (89\%) &    $0.1 \pm 0.1$ & $0.8 \pm 1.1$ & 73   & $1.7 \pm 5.2$ & $0.6 \pm 0.5$ \\
& \SpecThree    & 96 (92\%) &	 $0.1 \pm 0.2$ & $1.0 \pm 1.4$ & 59   & $1.2 \pm 1.0$ & $0.5 \pm 0.4$ \\
& \SpecFour    & 19 (79\%)   &	 $0.2 \pm 0.2$ & $1.6 \pm 3.1$ &  2    & $0.9, 0.8$ & $0.4, 1.0$ \\
& \SpecFive    & 38 (79\%)   &    $0.2 \pm 0.2$ & $1.4 \pm 1.8$ & 27   & $1.3 \pm 1.2$ & $0.4 \pm 0.4$ \\
& \SpecSix    & 40 (73\%)   &	 $0.1 \pm 0.1$ & $0.7 \pm 0.9$ & 20   & $0.8 \pm 0.8$ & $0.6 \pm 0.7$ \\
& \SpecOne    & 7 (29\%)   &	 $0.1\pm 0.1$ & $0.8 \pm 1.0$ & 2     & $1.3, 0.6$ & $0.4, 0.1$ \\
& All  & 299 (82\%) &	 $0.1 \pm 0.2$ & $1.0 \pm 1.5$ & 183 & $1.4 \pm 3.4$ & $0.5 \pm 0.5$ \\
\noalign{\smallskip}\hline\noalign{\smallskip}
\multirow{7}{*}{\rotatebox{90}{3: Combined}}
& \SpecTwo    & 101 (91\%) &    $0.1 \pm 0.1$ & $1.0 \pm 1.5$ & 73   & $1.8 \pm 5.2$ & $0.6 \pm 0.5$ \\
& \SpecThree    & 99 (95\%) &	 $0.2 \pm 0.2$ & $1.4 \pm 1.7$ & 61   & $1.3 \pm 1.0$ & $0.7 \pm 0.8$ \\
& \SpecFour    & 18 (75\%)   &	 $0.2 \pm 0.2$ & $2.8 \pm 3.4$ &  2    & $1.1, 1.1$ & $1.0, 1.3$ \\
& \SpecFive    & 41 (85\%)   &    $0.2 \pm 0.2$ & $2.1 \pm 2.9$ & 29   & $1.6 \pm 1.3$ & $0.6 \pm 1.0$ \\
& \SpecSix    & 47 (85\%)   &	 $0.1 \pm 0.1$ & $0.9 \pm 1.2$ & 24   & $0.8 \pm 0.8$ & $0.5 \pm 0.6$ \\
& \SpecOne    & 7 (29\%)   &	 $0.3\pm 0.3$ & $3.0 \pm 3.2$ & 3     & $1.0 \pm 0.7$ & $0.3 \pm 0.2$ \\
& All  & 313 (86\%) &	 $0.2 \pm 0.2$ & $1.4 \pm 2.0$ & 192 & $1.5 \pm 3.3$ & $0.6 \pm 0.7$ \\
\noalign{\smallskip}\hline
\end{tabular}
\end{table*}
\fi
\subsection{Intraoperative Registration}\label{sec:results_intraop_regi}
Using estimated segmentations and landmark locations of the 366 test images, each intraoperative registration strategy was run and compared to the ground truth pose estimates from the training dataset.
Registrations with pelvis rotation error less than $1\degree$ were defined as successful.
This error was computed in the projective frame with center of rotation at the ground truth midpoint between the two femoral heads.
Femur errors were computed using relative poses of the femur with respect to the pelvis in the APP with center of rotation at the ipsilateral femoral head.
Table~\ref{tab:results_regi_errors_all_specs} lists the mean registration errors for each left-out specimen and the errors over all specimens.
Depth estimation inaccuracies accounted for nearly all of the pelvis translation error, with mean errors about the X, Y, and Z axes of $0.1 \pm 0.1$ mm, $0.1 \pm 0.1$ mm and $1.4 \pm 2.0$ mm, respectively for method~3.
For each specimen, the decompositions of method~3's pelvis errors are listed in \SuppTab{\ref{tab:results_decomp_pelvis_regi_errors}}.

Two-tailed Mann-Whitney U tests were used to compare the magnitudes of the rotation and translation errors between methods 2 and 3.
Using a $0.005$ threshold, a significant difference was found between the pelvis rotation errors ($p<0.001$),
while no significant differences were found between pelvis translation errors ($p=0.045$), femur rotation errors ($p=0.089$), and femur translation errors ($p=0.268$).

Correlation coefficients between dice scores and the pelvis rotation and translation errors were calculated using Spearman's rank correlation coefficient.
For\\
method~2, the correlation coefficients for the pelvis rotation and translation errors were $-0.32$ and $-0.29$, respectively, and $-0.31$ and $-0.33$, respectively for\\
method~3.
The average of dice scores from the segmentations of hemipelves and femurs was used for this calculation.
%

The mean runtime for method~3 was $7.2 \pm 0.7$ seconds using a NVIDIA Tesla P100 (PCIe) GPU.
Examples of automatic annotation and registration with method~3 are shown in the supplementary video\footnote{\href{https://youtu.be/5AwGlNkcp9o}{https://youtu.be/5AwGlNkcp9o}}.
%
\begin{figure*}
\begin{centering}
\includegraphics[width=0.7\textwidth]{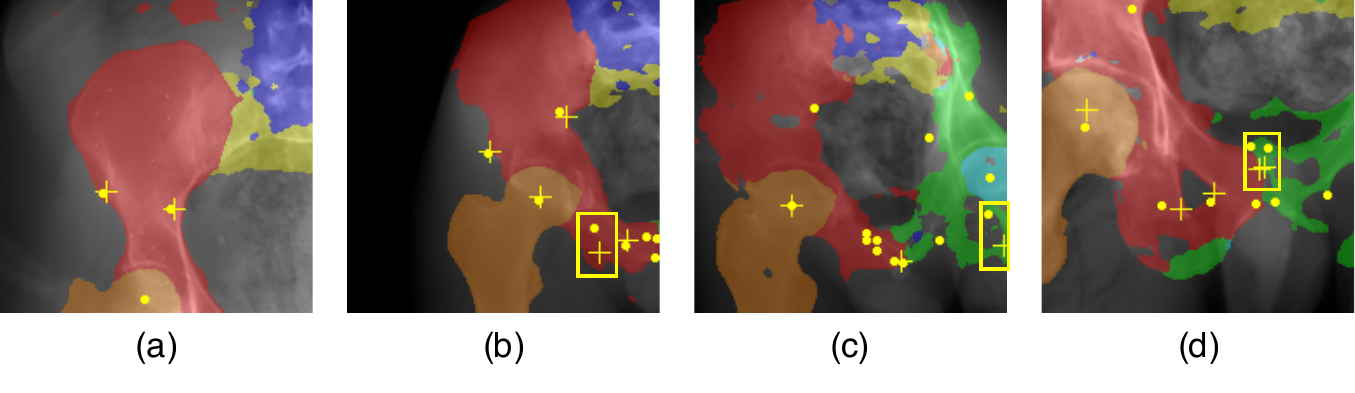}
\caption{Abnormal cases with C-arm poses different from the training dataset.
			 The lower hip is not visible in (a), however 2 landmarks were accurately detected allowing a successful pelvis registration.
			 Excessive pelvic tilt is shown in (b), (c), and (d) shows large magnification.
			 Detections with large errors are highlighted by yellow boxes.
			 In (b) a single landmark, out of five detections, had large error, which allowed the registration strategy to succeed. 
			 The C-arm pose of (c) causes the boundary along the left femoral neck to appear similar to that adjacent to the IOF in an AP view, resulting in a detection with large error.
			 Pelvis registration in (d) fails due to the large localization errors in each landmark.}
\label{fig:abnormal_cases_example}
\end{centering}
\end{figure*}
\vspace{-4mm}
\section{Discussion and Conclusion}\label{sec:discussion}
%
%
The naive intraoperative registration performed poorly, succeeding in only 18\% of trials, while the methods leveraging CNN annotations succeeded over 4 times as often.
Despite the fairly large false-negative detection rate of 17\%, an average of 7 landmarks per image were detected, allowing methods 2 and 3 to perform well.
Method~3's performance was robust when only 2, 3, and 4 landmarks were detected; reporting success in 2, 15, and 30 cases, respectively.
\Fig~\ref{fig:abnormal_cases_example} (a) shows an image with 2 detected landmarks and was registered successfully.
Highlighting the robustness gained from mixing intensity-features with landmark features, method~2 was only successful with these number of detections in 0, 7, and 26 cases, respectively.
The low false positive detection rate ensured that inconsistent features would not confound the registration.

Although the naive registration only succeeded in 66 cases, the mean femur rotation errors were about 1\degree~smaller than those of methods 2 and 3.
However, methods 2 and 3 were also successful in 62 and 64 of method~1's successes, each with a mean femur rotation error of $0.8\degree \pm 0.5 \degree$.
This indicates that the three methods perform comparably on images for which the naive approach succeeds.
Moreover, the larger errors of methods 2 and 3 in the remaining cases are in part caused by the more challenging pelvis registration problems presented in these images, for which the naive method was unable to solve.

Method~3 was robust to poor initializations, which most likely resulted in the larger number of successful pelvis registrations compared to those resulting from method~2.
In contrast to method~2, the objective function of method~3 never places penalties on the offsets of poses from their initial estimates.
The registration is free to minimize the image similarity term on the condition that known 3D landmarks project to approximately the correct location in 2D.
Conversely, the registration is free to minimize landmark reprojection error, so long as the candidate poses produce images that approximately match the observed image.
This is contrary to the standard approach for regularization used by method~2, which imposes limits on the amount of rigid movement, even when the initial estimates are far from the true poses.

Despite the significant difference in pelvis rotation errors between methods 2 and 3, we believe that the small error magnitudes resulting from both methods should not negatively impact the clinical application of either approach.

Table~\ref{tab:results_regi_errors_all_specs} shows that the mean and standard deviation of method~3's pelvis translation errors were both larger than those of method~2 by approximately $0.5$ mm.
It is possible that some cases of inaccurate landmark point estimates may have limited the influence of image similarities, resulting in the larger errors for method~3.
We believe that this issue may be overcome by replacing the landmark reprojection distances of~\eqref{eq:intraop_landmark_regularization} with the heatmap values at each projected 3D landmark.
Since the heatmaps encode landmark localization uncertainties, this modification should reduce the penalty of reprojection distances for inaccurately estimated landmarks.

In comparison to the femur, the pelvis is a larger, more complex shape, which occupies larger regions of the fluoroscopic views.
The pelvis' fluoroscopic appearance is thus more sensitive to rotational changes than that of the femur, which is consistent with the smaller rotation errors observed with registrations of the pelvis and shown in Table~\ref{tab:results_regi_errors_all_specs}.

For the intact hip anatomy that is considered in this paper, the connective tissues joining each femoral head to the acetabular regions of the pelvis cause each femur and the pelvis to mostly translate together.
As a result, the relative pose of a femur with respect to the pelvis contains very little translation.
This prior knowledge is incorporated into the registration strategies and causes mostly small femoral translations to be reported, which results in the small translation errors for femur registrations listed in Table~\ref{tab:results_regi_errors_all_specs}.
The influence of pelvis translation errors on these estimates was minimal, since nearly all of the pelvis translation error was found in the depth direction, which has a minor impact on single-view appearance.
%
%
%
%
%
\begin{figure*}
\begin{centering}
\includegraphics[width=0.9\textwidth]{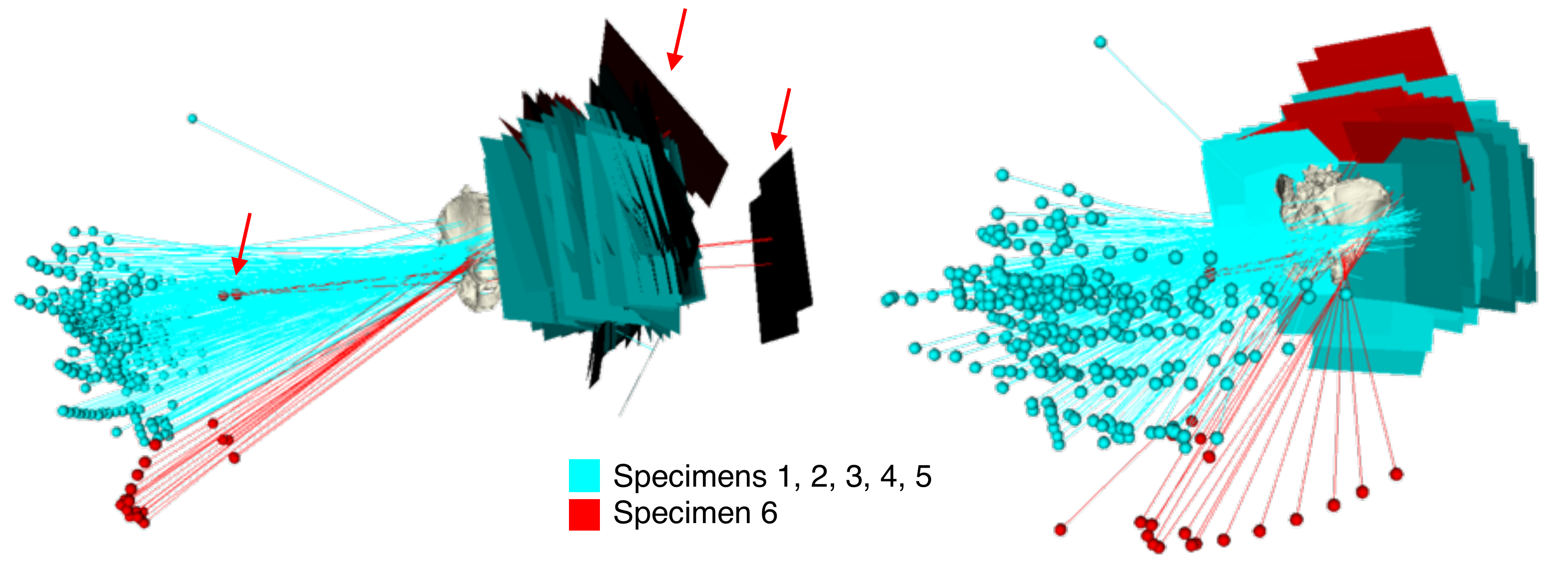}
\caption{A visualization of all ground truth projection geometries using the APP as the world coordinate frame.
		     Each sphere represents a position of the X-ray source, each square represents the position of the X-ray detector, and each line connects the X-ray source to the principal point on the detector.
		     Red arrows highlight difficult to see geometries of specimen~\SpecOne.
		     Most of the poses are contained within a $60\degree$ range of C-arm orbital rotations and a $30\degree$ range of pelvic tilts. 
			 }
\label{fig:pat_1_geoms}
\end{centering}
\end{figure*}

The performance of our approach degrades as images collected with C-arm poses not found in the training data set are processed.
This is highlighted in Table~\ref{tab:results_regi_errors_all_specs}, showing the poor performance of specimen~\SpecOne~compared to all other specimens.
When testing on spec.~\SpecOne, average landmark localization error was $10.0$ mm, with a false negative rate of $30\%$, and successful pelvis registration rate of $29\%$ for both methods 2 and 3.
\Fig~\ref{fig:pat_1_geoms} shows a visualization of all 366 ground truth projection geometries.
The geometries associated with spec.~\SpecOne~are clearly collected at different C-arm poses than those used for training the network tested on spec.~\SpecOne.
Three examples of spec.~\SpecOne~ are shown in \Fig~\ref{fig:abnormal_cases_example} (b)-(d).
This limitation may be overcome by collecting more fluoroscopy data for training.
However, it is conceivable that some C-arm poses encountered during testing will still be absent during training.
By augmenting actual fluoroscopy with realistic synthetic fluoroscopy~\cite{unberath2019enabling} during training, we believe that quality performance at these ``missing'' poses may be achieved.

Large variations in dice scores may result from the slight mislabeling of a narrow structure, such as the ilium in some views.
These small segmentation errors are not expected to negatively impact registration performance, as the estimated labels are used to weight the contribution of local, overlapping, patches to the image similarity term.
Therefore, it is not surprising that a weak correlation between dice scores and registration errors was indicated by the Spearman rank coefficient values.

The automated annotation and registration techniques proposed in this paper could streamline intraoperative workflows related to intact hip anatomy, such as osteotomy planning~\cite{gottschling2005intraoperative}, robotic drilling~\cite{gao2020fiducial}, and 3D reconstructions of bone~\cite{reyneke2018review} or implanted tracking fiducials~\cite{grupp2019fast}.
Extending the annotation component to label additional objects, such as surgical instruments, artificial implants, and bone fragments could enable automatic registration of surgically modified hip anatomy and is the subject of future research.
Although all possible patches are currently evaluated when computing image similarities,
the semantic labeling of fluoroscopy should enable much smaller subsets of anatomically relevant patches to be used during the registration.
By only rendering the DRR pixels which intersect this subset of patches, significant reductions of registration runtimes may be possible.

In conclusion, this paper has demonstrated that small annotated datasets of actual hip fluoroscopy may be used to train CNN models capable of state-of-the-art segmentation and landmark localization results.
Furthermore, we have shown that by coupling the automatic annotations produced by the CNN models with the image intensities used during 2D/3D registrations, robustness against poor initializations is possible.
This is a clinically relevant result, as this robustness removes the need for manual initialization and allows navigation to be more naturally integrated into existing surgical workflows.
To our knowledge, the dataset presented in this paper is the first annotated dataset of \textit{actual} hip fluoroscopy, consisting of individual bone segmentations and anatomical landmark locations.
We have also made this dataset publicly available.\footnote{\href{https://github.com/rg2/DeepFluoroLabeling-IPCAI2020}{https://github.com/rg2/DeepFluoroLabeling-IPCAI2020}}
Creation of the precise labels found in the training dataset was made possible by extending existing 2D/3D registration technology into a new \textit{offline} and semi-automatic annotation pipeline. 
We believe this ground truth labeling method will translate to fluoroscopy of other anatomy and enable machine learning applications in other specialties.
%
%
\begin{acknowledgements}
We thank Mr. Demetries Boston for assisting with the cadaveric data acquisition.
This research was supported by NIH/NIBIB grants R01EB006839,\\
R21EB020113, Johns Hopkins University Internal Funds, and a Johns Hopkins University Applied Physics Laboratory Graduate Student Fellowship.
Part of this research project was conducted using computational resources at the Maryland Advanced Research Computing Center (MARCC).
\end{acknowledgements}
{\small
\textbf{Compliance with Ethical Standards}\\
\textbf{Conflict of Interest:} The authors declare that they have no conflict of interest.\\
\textbf{Ethics Approval:} This article does not contain any studies with human participants performed by any of the authors.\\
\textbf{Informed Consent:} This article does not contain patient data.
}
%
%
%
%
%
%
\bibliographystyle{spmpsci}      
\bibliography{IEEEabrv,refs}  
%
\appendix
\renewcommand{\thetable}{S-\arabic{table}}
\renewcommand{\thefigure}{S-\arabic{figure}}
\setcounter{figure}{0}
\setcounter{table}{0}
\section{Supplementary Methods}\label{sec:supp_methods}
%
%

%
%
\subsection{2D/3D Registration}\label{sec:supp_methods_regi_2d3d}
The $\mathfrak{se}(3)$ Lie algebra parameterization of $SE(3)$ with reference point at the initial pose estimate of the object, is used during the optimization of rigid poses.
The $\mathfrak{so}(3)$ Lie algebra parameterization of $SO(3)$ is used when only optimizing over the rotation component.
Optimization is performed with respect to the projective frame when performing registrations of the pelvis only.
When registering each individual femur or all objects simultaneously, optimization is run with respect to the anterior pelvic plane (APP).
\subsection{Training Dataset Creation}\label{sec:supp_methods_dataset_creation}
\begin{figure*}
\begin{centering}
\includegraphics[width=0.75\textwidth]{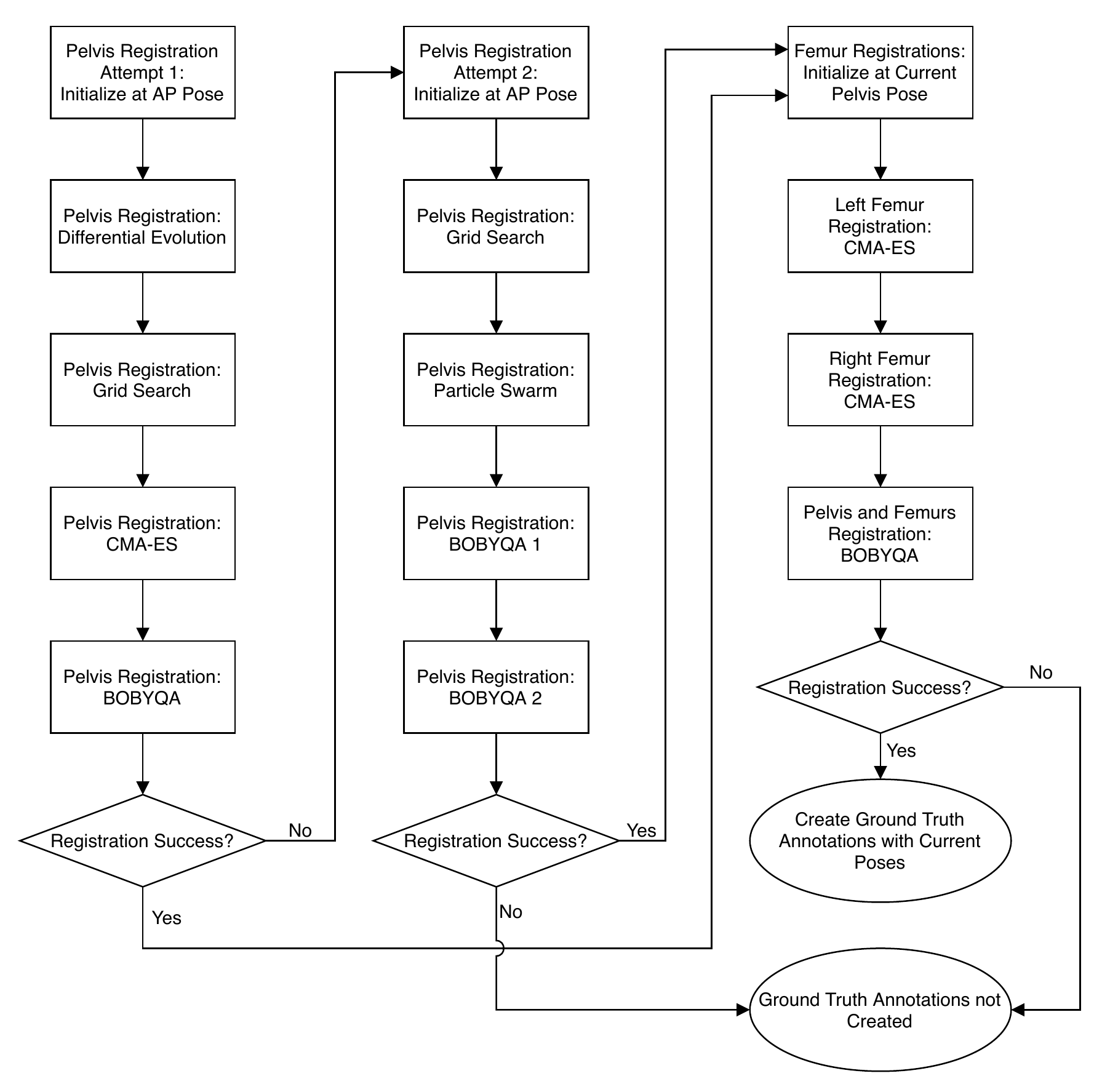}
\caption{High-level workflow of the registrations used for ground truth labeling of fluoroscopy.}
\label{fig:ground_truth_regi_workflow}
\end{centering}
\end{figure*}

\Fig~\ref{fig:ground_truth_regi_workflow} shows a high-level workflow of the registrations used during creation of the training data set.
The amounts of downsampling used for each method are listed in Table~\ref{tab:supp_ds_factors}.
Any box constraints used by the following methods are listed in Table~\ref{tab:supp_se3_box_constraints}.
\subsubsection{Computationally Expensive Automatic Pelvis Registration}
Two attempts are made to solve for the pose of the entire pelvis.
The first attempt sequentially applies the following optimization strategies: Differential Evolution (DE)~\cite{storn1997differential}, exhaustive grid search, Covariance Matrix Adaptation: Evolutionary Search (CMA-ES)~\cite{hansen2001completely}, and Bounded Optimization by Quadratic Approximation (BOBYQA)~\cite{powell2009bobyqa}.

The DE optimization uses a regularizer designed to penalize poses which:
do not project at least one femoral head center within the 2D image bounds,
project inferior landmarks above superior landmarks in the image,
or place either femoral head center behind the detector or too close to the X-ray source.
This regularization is defined by $\mathcal{R}_\text{DE}$ in \eqref{eq:de_reg_main_def}.
{
\begin{equation} \label{eq:de_reg_main_def}
\begin{aligned}
	\mathcal{R}_\text{DE} \left( \theta_P \right) =& 2 \left[ \mathcal{R}_\text{visible} \left( p_\text{FH}^\text{left} ; \theta_P \right) \mathcal{R}_\text{visible} \left( p_\text{FH}^\text{right} ; \theta_P \right) \right] + \\
														& 2 \left[ \mathcal{R}_\text{depth} \left( p_\text{FH}^\text{left} ; \theta_P \right) + \mathcal{R}_\text{depth} \left( p_\text{FH}^\text{right} ; \theta_P \right) \right] + \\
														& \left[ \mathcal{R}_\text{up} \left( p_\text{ASIS}^\text{left}, p_\text{IOF}^\text{left} ; \theta_P \right) + \mathcal{R}_\text{up} \left( p_\text{ASIS}^\text{right}, p_\text{IOF}^\text{right} ; \theta_P \right) \right]
\end{aligned}
\end{equation}
}
The individual penalty applied for projecting a point outside of the field of view is defined in \eqref{eq:de_reg_visible}.
The number of pixels, in the row direction, by which the point is projected ``out-of-bounds'' is indicated by $r$, and $c$ is the corresponding value in the column direction.
Both $r$ and $c$ are zero-valued for projected locations within the image bounds.
{
\begin{equation} \label{eq:de_reg_visible}
	\mathcal{R}_\text{visible} \left( p ; \theta_P \right) = r^2\left(p ; \theta_P \right) + c^2 \left( p ; \theta_P \right)
\end{equation}
}
The individual penalty applied for points at unexpected depths is shown in \eqref{eq:de_reg_depth}.
The depth of a point, as a ratio of source-to-detector depth, is denoted by $d$.
Zero indicates a depth equal to the X-ray source and one indicates the depth of the X-ray detector.
{
\begin{equation} \label{eq:de_reg_depth}
	\mathcal{R}_\text{depth} \left( p ; \theta_P \right)  = \begin{cases}
								d \left( p ; \theta_P \right)^2 & \text{if } d \left( p ; \theta_P \right) \geq 1 \\
								100 \left[ 0.7 - d \left( p ; \theta_P \right) \right]^2 & \text{if } d \left( p ; \theta_P \right) \leq 0.7 \\
								0 & \text{otherwise}
							\end{cases} 
\end{equation}
}
The individual penalty applied for projecting a certain point ``above'' another is defined in \eqref{eq:de_reg_up}.
For image visualization in this paper, smaller row values are located above larger values.
Each image is assumed to be oriented ``patient-up,'' so that superior regions occupy smaller row locations than inferior regions.
Therefore, $\mathcal{R}_\text{up} \left( p_\text{ASIS}^\text{left}, p_\text{IOF}^\text{left} ; \theta_P \right)$ applies a penalty when the, relatively inferior, IOF landmark is projected above the, relatively superior, ASIS landmark.
{
\begin{equation} \label{eq:de_reg_up}
	\mathcal{R}_\text{up} \left( p, q ; \theta_P \right) =
		\begin{cases}
	 		\left( \mathcal{P} \left( q ; \theta_P \right)_\text{row} - \mathcal{P} \left( p_ ; \theta_P \right)_\text{row} \right)^2 \qquad \\
						\hfill \text{if } \mathcal{P} \left( q ; \theta_P \right)_\text{row}  < \mathcal{P} \left( p_ ; \theta_P \right)_\text{row} \\ \\
			0 \hfill \text{otherwise}
		\end{cases}
\end{equation}
}

%
DE is run for 400 iterations, with a population size of 1000, and a cross-over probability of $CR = 0.2$.
Dithering is used to choose the evolution rate parameter, $F \sim U(0.5,1)$, for each mutation vector.

The grid search is performed over a smaller region than the DE search and does not use regularization.
Table~\ref{tab:supp_se3_grid_incs} lists the grid search increments used.
After grid search the same strategy used in \cite{grupp2019pose} for registering the pelvis in a single view is applied.
CMA-ES uses a population size of 100 and regularizes the current pose according to its Euler-decomposition in the projective frame.
The decomposed values are assumed to be independent and drawn from $N(0,\sigma_i)$, for\\
$\sigma_i = \{ 10\degree, 10\degree, 10\degree, 20, 20, 100 \}$.
Table~\ref{tab:supp_cmaes_params} lists the CMA-ES parameters.

If the first pelvis registration attempt is not successful, then another attempt is made using the following sequence of optimizations: exhaustive grid search, Particle Swarm Optimization (PSO)~\cite{shi1998modified}, and two runs of BOBYQA at increasing resolutions levels.
No regularization is used during this attempt.
The grid search used during this attempt is performed at coarser increments, but over a larger region, compared to the first attempt's grid search.
PSO was run for 50 iterations, with $21,000$ particles, momentum $\omega = 0.7298$, local weight upper bound $\varphi_p = 1.4961$, and global weight upper bound $\varphi_g = 1.4961$.

No further attempt is made to annotate the current fluoroscopy image if this attempt is also unsuccessful.
\subsubsection{Registration of the Femurs}
If the pelvis registration is successful, then an attempt is made to register the left and right femurs.
This strategy first registers the left femur only, keeping the pelvis fixed at its current pose estimate.
Next, the right femur is registered, again keeping the pelvis fixed.
Both of these registrations use CMA-ES.
Contrary to the previous registrations, these only search the 3D space of rotations, with the center of rotation fixed at the ipsilateral femoral head center.
Regularization is applied to the total rotation magnitude using a folded normal distribution with $\mu = 45 \degree$ and $\sigma = 45 \degree$.
Table~\ref{tab:supp_cmaes_params} lists the CMA-ES parameters.
Once again, successful registrations of each object are manually verified.
\begin{table}
\centering
\caption{Amount of downsampling along each 2D image dimension applied during each optimization.}
\label{tab:supp_ds_factors}       
\begin{tabular}{c l r}
\hline\noalign{\smallskip}
Object & Strategy & Factor \\
\noalign{\smallskip}\hline\noalign{\smallskip}
\multirow{4}{*}{Pelvis Attempt 1}
& DE               & $32\times$ \\
& Grid             & $32\times$   \\
& CMA-ES      & $8\times$   \\
& BOBYQA     & $4\times$   \\ \noalign{\smallskip}\hline \noalign{\smallskip}
\multirow{4}{*}{Pelvis Attempt 2}
& Grid             & $32\times$ \\
& PSO            & $32\times$   \\
& BOBYQA 1 & $8\times$   \\
& BOBYQA 2 & $4\times$   \\ \noalign{\smallskip}\hline \noalign{\smallskip}
\multirow{1}{*}{Femurs}
& CMA-ES     & $8\times$ \\ \noalign{\smallskip}\hline \noalign{\smallskip}
\multirow{1}{*}{All Objects}
& BOBYQA    & $4\times$ \\
\noalign{\smallskip}\hline
\end{tabular}
\end{table}
\begin{table*}
\centering
\caption{The $\mathfrak{se}(3)$ box constraints used for the registrations used to obtain ground truth annotations.
			 For the all objects case, the box constraints are repeated for the three objects.}
\label{tab:supp_se3_box_constraints}       
\begin{tabular}{c l r r r r r r}
\hline\noalign{\smallskip}
\multirow{2}{*}{Object} & \multirow{2}{*}{Strategy} &  \multicolumn{6}{c}{Dimension} \\
 & & 1 & 2 & 3 & 4 & 5 & 6 \\
\noalign{\smallskip}\hline\noalign{\smallskip}
\multirow{3}{*}{Pelvis Attempt 1}
& DE               & $\pm 60\degree$  & $\pm 40\degree$  & $\pm 10\degree$  & $\pm 200$  & $\pm 200$ & $\pm 250$ \\
& Grid             & $\pm 5\degree$    & $\pm 5\degree$    & $\pm 1\degree$    & $\pm 10$    & $\pm 10$   & $\pm 50$   \\
& BOBYQA     & $\pm 2.5\degree$ & $\pm 2.5\degree$ & $\pm 2.5\degree$ & $\pm 5$      & $\pm 5$     & $\pm 10$   \\ \noalign{\smallskip}\hline \noalign{\smallskip}
\multirow{4}{*}{Pelvis Attempt 2}
& Grid             & $\pm 60\degree$  & $\pm 40\degree$  & $0\degree$           & $\pm 200$  & $\pm 200$ & $\pm 250$ \\
& PSO            & $\pm 7.5\degree$ & $\pm 10\degree$  & $\pm 10\degree$  & $\pm 20$    & $\pm 20$   & $\pm 25$   \\
& BOBYQA 1 & $\pm 5\degree$    & $\pm 5\degree$    & $\pm 5\degree$     & $\pm 10$    & $\pm 10$   & $\pm 20$   \\
& BOBYQA 2 & $\pm 2.5\degree$ & $\pm 2.5\degree$ & $\pm 2.5\degree$  & $\pm 5$      & $\pm 5$     & $\pm 10$   \\ \noalign{\smallskip}\hline \noalign{\smallskip}
\multirow{1}{*}{All Objects}
& BOBYQA    & $\pm 2.5\degree$ & $\pm 2.5\degree$ & $\pm 2.5\degree$  & $\pm 2.5$  & $\pm 2.5$   & $\pm 2.5$ \\
\noalign{\smallskip}\hline
\end{tabular}
\end{table*}
\begin{table}
\centering
\caption{The $\mathfrak{se}(3)$ increments used for each grid search.}
\label{tab:supp_se3_grid_incs}       
\begin{tabular}{c r r r r r r}
\hline\noalign{\smallskip}
\multirow{2}{*}{Pelvis Attempt} & \multicolumn{6}{c}{Dimension} \\
& 1 & 2 & 3 & 4 & 5 & 6 \\
\noalign{\smallskip}\hline\noalign{\smallskip}
1 & $1 \degree$    &  $1 \degree$ &  $1 \degree$ &  2 &   2  & 10 \\
2 & $7.5 \degree$ &  $5 \degree$ &  $0 \degree$ & 20 & 20 & 25 \\
\noalign{\smallskip}\hline
\end{tabular}
\end{table}
\begin{table}
\centering
\caption{CMA-ES population size and initial $\sigma$ parameters.}
\label{tab:supp_cmaes_params}       
\begin{tabular}{c r r r r r r r}
\hline\noalign{\smallskip}
\multirow{2}{*}{Object} & \multirow{2}{*}{Pop. Size} & \multicolumn{6}{c}{Dimension} \\
 & & 1 & 2 & 3 & 4 & 5 & 6 \\
\noalign{\smallskip}\hline\noalign{\smallskip}
Pelvis &  100 & $15\degree$ & $15\degree$ & $30\degree$ &  50 &  50 & 100 \\
Femur & 100 & $30\degree$ & $25\degree$ & $15\degree$ & --    & --   & -- \\
\noalign{\smallskip}\hline
\end{tabular}
\end{table}
\subsection{Network Architecture and Training}\label{sec:supp_methods_network}
\subsubsection{Architecture}
Fluoroscopy data is downsampled $8\times$ from $1436\times1436$ pixels, after border cropping, to $180\times180$ pixels.
Each image is padded to $192\times192$ using reflection.
This is necessary in order to obtain output segmentations and heatmaps at $180\times180$ after several rounds of U-Net downsampling and upsampling.

\begin{figure}
\begin{centering}
\includegraphics[width=\columnwidth]{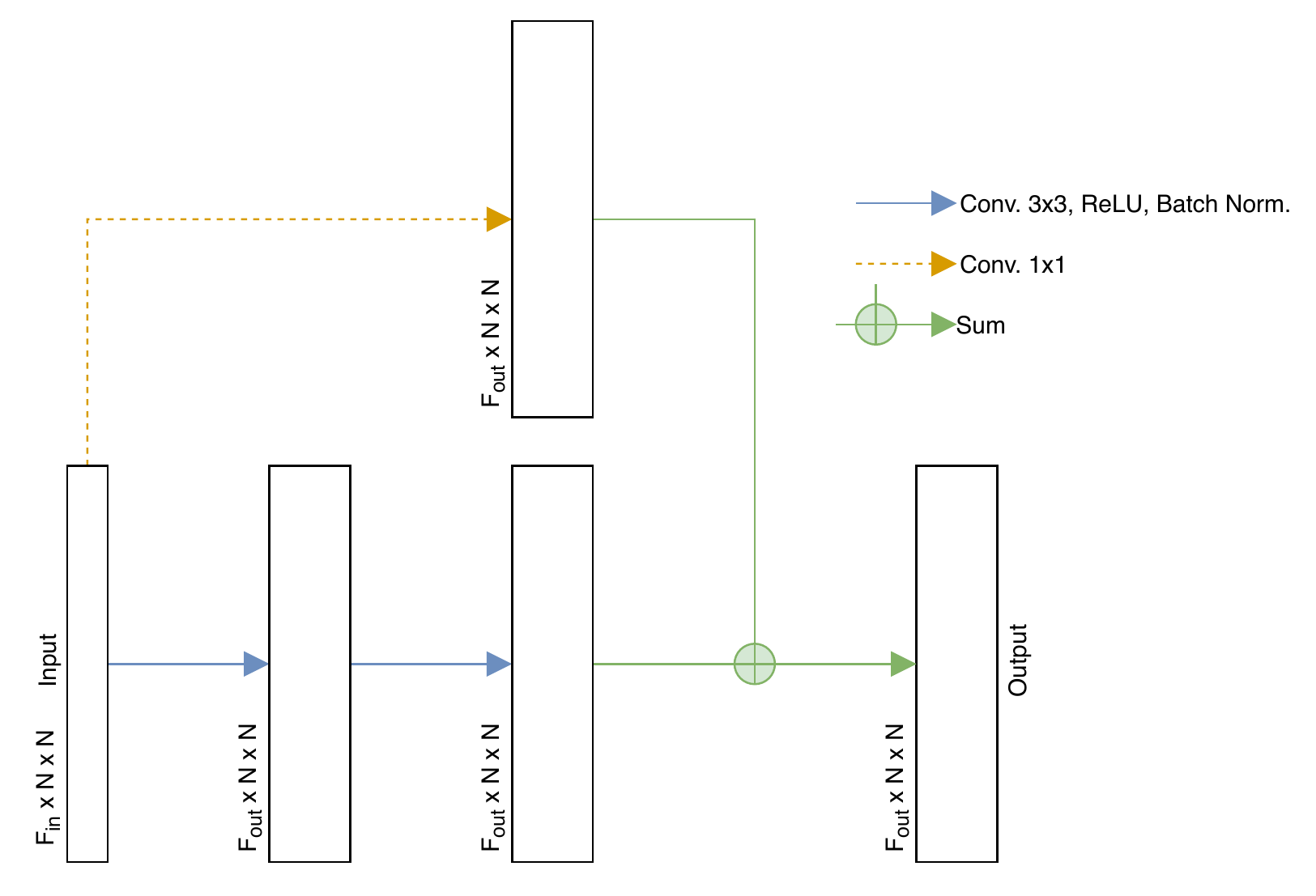}
\caption{The architecture of an individual U-Net block used in this work.}
\label{fig:unet_block}
\end{centering}
\end{figure}

\begin{figure*}
\begin{centering}
\includegraphics[width=0.9\textwidth]{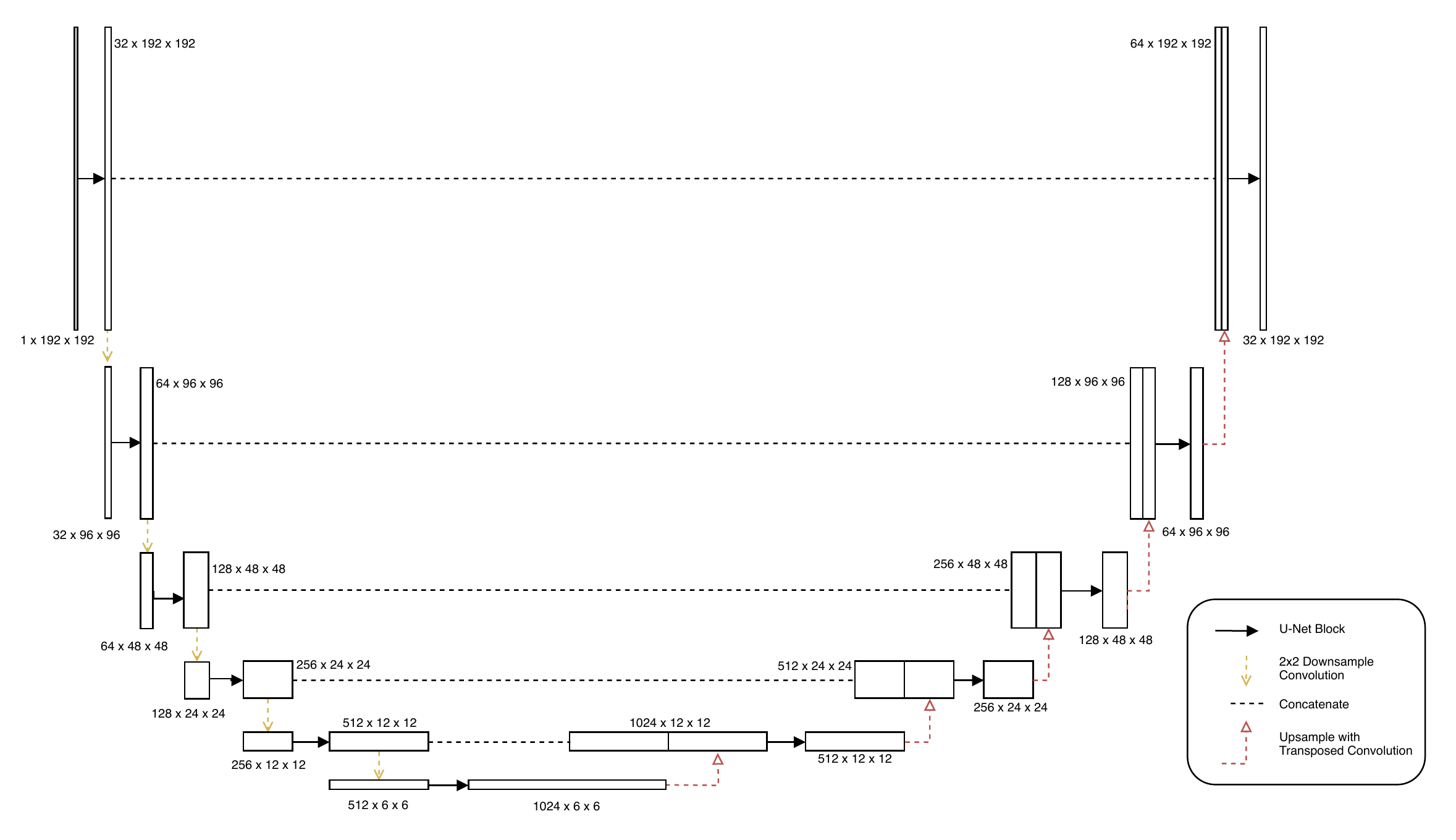}
\caption{The architecture of the U-Net encoder-decoder used in this work.}
\label{fig:unet_enc_dec}
\end{centering}
\end{figure*}

\begin{figure}
\begin{centering}
\includegraphics[width=\columnwidth]{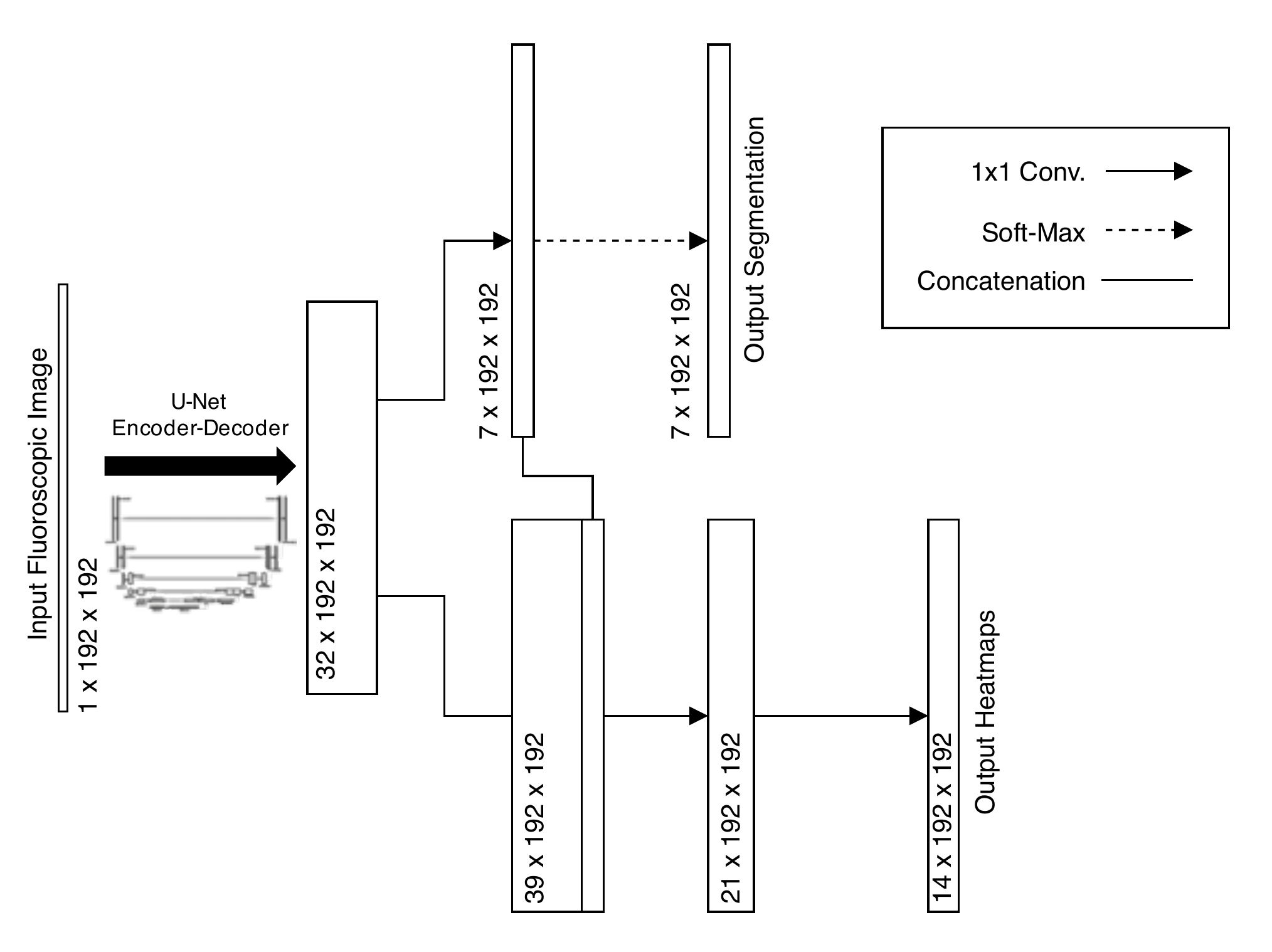}
\caption{High-level network structure used in this paper.
			 After an image is processed through a U-Net encoder-decoder module, the segmentation is computed using the standard approach.
			 Segmentation features prior to soft-max are concatenated with the features output from the encoder-decoder and two 1x1 convolutions are used to estimate the landmark heatmaps.}
\label{fig:network}
\end{centering}
\end{figure}

\Fig~\ref{fig:unet_block} shows the architecture of an individual U-Net block.
Each U-Net block consists of two consecutive sequences of: a 3x3 convolution, a ReLU non-linear activation, and batch normalization~\cite{ioffe2015batch}.
Residual connections \cite{he2016deep} are also applied in each block.
Zero padding is used for all convolutions.
The entire U-Net encoder-decoder is shown in \Fig~\ref{fig:unet_enc_dec} and a high-level diagram of the entire network is shown in \Fig~\ref{fig:network}.
\subsubsection{Loss Functions}
For the segmentation branch of the network, the differentiable dice score \cite{milletari2016v} is computed for each class and then averaged as shown in \eqref{eq:dice_loss}.
$N_C$ is equal to the number of classes including background (7 in this paper), $w$ are the network weights, $\widehat{M}^{(k)}$ is the ground truth segmentation mask for class $k$, and $M^{(k)}$ is the estimated segmentation mask for class $k$.
\begin{equation} \label{eq:dice_loss}
	D\left( w \right) = \frac{1}{N_C} \sum_{k=1}^{N_C}
		\frac{ 2 \displaystyle\sum_{x,y} M^{(k)}(x,y; w) \widehat{M}^{(k)}(x,y) }
			   { \displaystyle\sum_{x,y} M^{(k)}(x,y; w)^2 +  \displaystyle\sum_{x,y}  \widehat{M}^{(k)}(x,y)^2 }
\end{equation}
%

Ground truth heatmaps for each landmark location,\\
$ ( \widehat{x}^{(l)}, \widehat{y}^{(l)} )$, are defined by \eqref{eq:gt_heatmap}, which is a symmetric 2D normal distribution with mean $ ( \widehat{x}^{(l)}, \widehat{y}^{(l)} )$ and $\sigma = 3.88$ mm ($2.5$ pixels).
\begin{equation} \label{eq:gt_heatmap}
	\widehat{h}^{(l)}(x,y) = 
		\begin{cases}
			\left( 2 \pi \sigma^2 \right)^{-1} \textrm{exp} \left\{ - \frac{ \left( x - \widehat{x}^{(l)} \right)^2 + \left( y - \widehat{y}^{(l)} \right)^2 } { 2 \sigma^2 }\right\} \qquad \\
			\hfill \mbox{if } ( \widehat{x}^{(l)}, \widehat{y}^{(l)} ) \mbox{ is visible} \\ \\
			0 \hfill \mbox{otherwise}
		\end{cases}	
\end{equation}

For two equal sized images $A$ and $B$, NCC is defined in \eqref{eq:ncc}.
Each image has $P$ pixels, means $\mu_A$ and $\mu_B$, and standard deviations $\sigma_A$ and $\sigma_B$.
\begin{equation} \label{eq:ncc}
	NCC\left( A, B \right) = \sum_{x,y} \frac{\left( A\left( x,y \right) - \mu_A \right) \left( B\left( x, y \right) - \mu_B \right)}{P \sigma_A \sigma_B}
\end{equation}
The average NCC value is computed over all estimated heatmaps, as shown in \eqref{eq:heatmap_loss}.
$N_L$ denotes the number of heatmaps/landmarks (14 in this paper), $\widehat{h}^{(l)}$ is the ground truth heatmap for landmark $l$ as defined in \eqref{eq:gt_heatmap}, and $h^{(l)}(w)$ is the estimated heatmap.
\begin{equation} \label{eq:heatmap_loss}
	H\left(w\right) = \frac{1}{N_L} \sum_{l=1}^{N_L} NCC \left( {h}^{(l)}(w),  \widehat{h}^{(l)} \right) 
\end{equation}

The dice and heatmap terms are combined into the final loss shown in \eqref{eq:total_loss}.
In order to weight the dice and heatmap terms equally, $H(w)$ is scaled and shifted to the range of $[0,1]$.
Since the optimization during training seeks to find a minimum, the combined term is negated.
\begin{equation} \label{eq:total_loss}
	\mathcal{L} \left( w \right) =  -\left[ D\left(w\right) + \frac{1}{2} \left( H\left(w\right) + 1 \right) \right]
\end{equation}
\subsubsection{Data Augmentation}
\begin{figure*}
\begin{centering}
\includegraphics[width=0.725\textwidth]{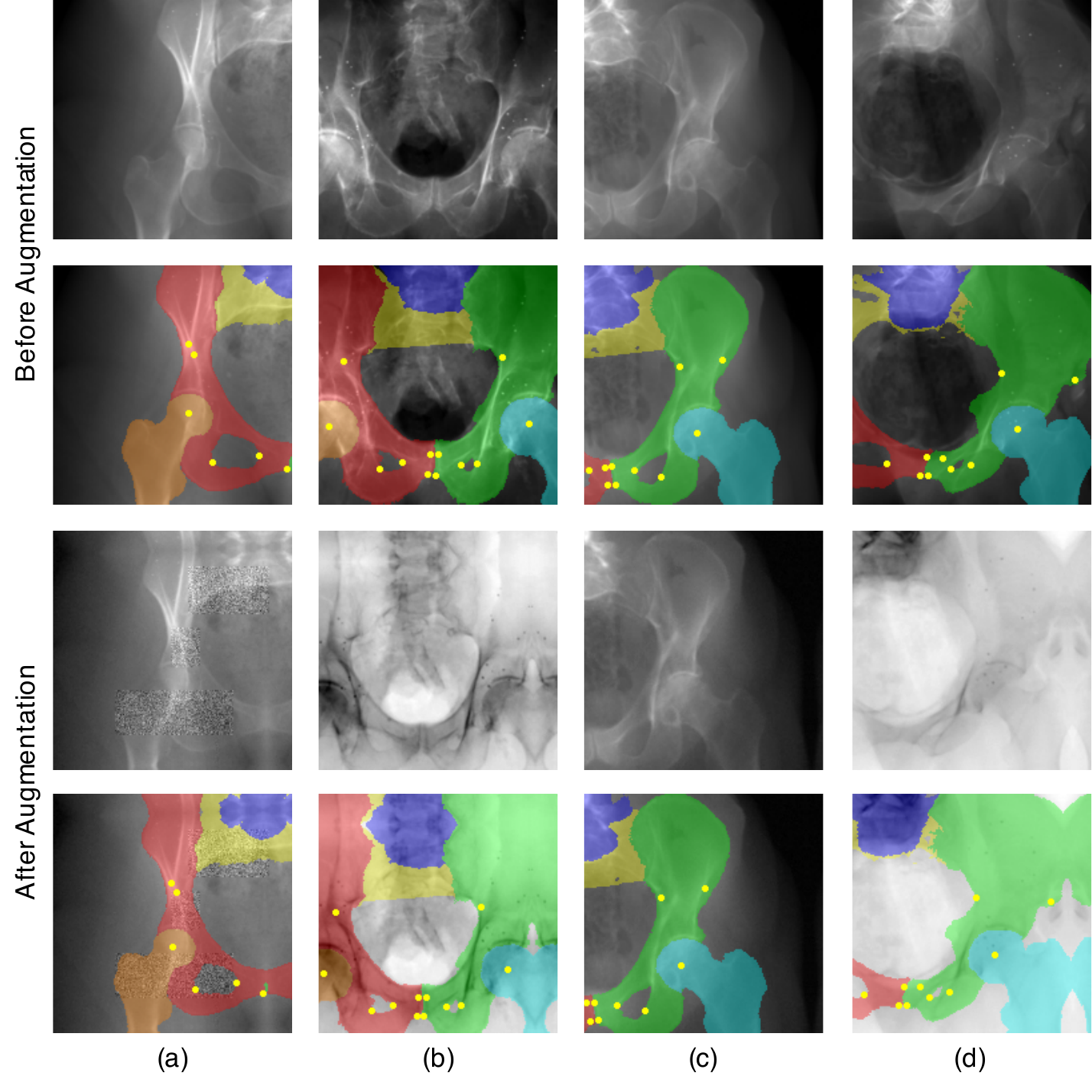}
\caption{Example data augmentation of the projections from \fig~\ref{fig:gt_and_est_segs_lands_example}. 
			 The original projections are shown in the top row, and shown again in the second row with the original annotations overlaid.
			 Projections after augmentation are shown in the third row, and the augmented annotations are overlaid in the bottom row.}
\label{fig:data_aug_example}
\end{centering}
\end{figure*}

Table~\ref{tab:data_aug} lists the operations performed when an image is randomly selected to be augmented during training.
Images are padded to $384 \times 384$ using reflection prior to warping, in order to avoid possible intensity discontinuities.
\Fig~\ref{fig:data_aug_example} shows data before, and after, augmentation. 
\begin{table}
\centering
\caption{Operations performed during data augmentation.}
\label{tab:data_aug}       
\begin{tabular}{ p{75pt} p{130pt} }
\hline\noalign{\smallskip}
Method & Description \\
\noalign{\smallskip}\hline\noalign{\smallskip}
Intensity Inversion & With probability $0.5$ \\
Additive Random Noise & $N(0, \sigma)$, $\sigma \sim U(0.005, 0.01)$ \\
Gamma Correction & $\gamma \sim U(0.7,1.3)$ \\
Affine Warp & Translation direction uniformly sampled \\
                    & Translation magnitude from $U(0,20)$ pixels \\
                    & Rotation angle from $U(-5\degree,+5\degree)$ \\
                    & Shear angle from $U(-2\degree, +2\degree)$ \\
                    & Scale from $U(0.9, 1.1)$ \\
Local Corruption & With probability $0.25$ \\
						   & Number of rectangular regions from $U(\{ 1, 2, 3, 4, 5 \})$ \\
						   & Region dimensions from $N(d, d)$, $d = 0.15 \times \text{image width}$ \\
						   & Location uniformly sampled, rejection sampling to ensure region is within image \\
                            & Additive noise from $N(0,0.2m)$, $m$ is the range of intensities in a region \\
\noalign{\smallskip}\hline
\end{tabular}
\end{table}
\subsection{Intraoperative Registration}\label{sec:supp_methods_intraop_regi}
For intraoperative method 2, using PnP initialization, regularization during CMA-ES pelvis registration is identically to that used when creating the training data set in ``Pelvis Attempt 1.''
For intraoperative method 3, combing intensity features and landmarks, a single landmark is used to recover translation when computing the initial AP pose.
Since any single landmark is not visible in all images, the following order of preference is used to select a landmark:
L.~FH,
R.~FH,
L.~IOF,
R.~IOF,
L.~IPS,
R.~IPS,
L.~MOF,
R.~MOF,
L.~SPS,
R.~SPS,
L.~GSN,
R.~GSN,
L.~ASIS,
R.~ASIS.
For regularization, $\sigma_\ell = 19.4$ mm.

During CMA-ES registration of the pelvis, $8\times$ downsampling is used along with the parameters listed in Table~\ref{tab:supp_cmaes_params}.
For BOBYQA registration of the pelvis, $4\times$ downsampling is used along with the BOBYQA box constraints for ``Pelvis Attempt 1'' in Table~\ref{tab:supp_se3_box_constraints}.
\section{Supplementary Results}\label{sec:supp_results}
\subsection{Annotated Dataset Creation}\label{sec:supp_results_dataset_creation}
Table~\ref{tab:training_data} lists the counts of the total number of images initially collected, the number of images with successful ground truth annotations, and the number of images with sufficient fields to view to perform femur registration.
Using a NVIDIA Tesla P100 (PCI-e), mean runtimes of $60.3 \pm 13.3$, $142.5 \pm 35.8$, and $2.5 \pm 0.3$ seconds were measured for attempt 1 of pelvis registration, attempt 2 of pelvis registration, and femur registration, respectively.
\begin{table}
\centering
\caption{The number of fluoroscopy images identified for potential use and the number of images used for network training.
		     Only images which were successfully registered with the ground truth labeling method were used for training.
		     Of the images used for training, counts of the images with sufficient visibility of the left and right femurs for registration purposes are also listed.
		     All specimens except one are used when training a specific network; the images for the left-out specimen are used as the test dataset.}
\label{tab:training_data}       
\begin{tabular}{p{28pt} p{31pt} p{37pt} p{43pt} p{43pt}}
\hline\noalign{\smallskip}
Specimen & \# Total Images & \# Images Used For Training & \# Training Images for L. Femur & \# Training Images for R. Femur \\
\noalign{\smallskip}\hline\noalign{\smallskip}
\SpecTwo    & 119 & 111 & 52 & 27\\
\SpecThree    & 108 & 104 & 39 & 24\\
\SpecFour    & 30 & 24 & 0 & 2\\
\SpecFive    & 53 & 48 & 17 & 18\\
\SpecSix    & 63 & 55 & 13 & 16\\
\SpecOne    & 26 & 24 & 0 & 12\\
All  & 399&366 & 121 & 99 \\
\noalign{\smallskip}\hline
\end{tabular}
\end{table}
\subsection{Segmentation and Landmark Localization}\label{sec:supp_results_seg_lands}
The mean training time for each network was $0.8 \pm 0.1$ hours using a NVIDIA Tesla P100 (PCI-e) GPU.

A listing of mean dice coefficients for each object of each ``left-out'' specimen is shown in Table~\ref{tab:results_dice_scores}.
\begin{table*}
\centering
\caption{Average dice coefficients obtained from each trained network from the leave-one-specimen-out experiment.
			 Actual dice coefficient is reported, not dice loss defined by \eqref{eq:dice_loss}.}
\label{tab:results_dice_scores}       
\begin{tabular}{r r r r r r r}
\hline\noalign{\smallskip}
\multirow{2}{*}{Specimen} & \multicolumn{6}{c}{Object Dice Coefficients} \\ \cline{2-7}
& L. Hemipelvis & R. Hemipelvis & L. Femur & R. Femur & Vertebrae & Sacrum \\
\noalign{\smallskip}\hline\noalign{\smallskip}
\SpecTwo    & $0.89 \pm 0.15$  & $0.89 \pm 0.13$  & $0.93 \pm 0.17$  & $0.78 \pm 0.37$  & $0.72 \pm 0.21$  & $0.63 \pm 0.09$     \\
\SpecThree    & $0.86 \pm 0.23$  & $0.85 \pm 0.22$  & $0.91 \pm 0.23$  & $0.94 \pm 0.18$  & $0.81 \pm 0.09$  & $0.66 \pm 0.12$     \\
\SpecFour    & $0.89 \pm 0.07$  & $0.91 \pm 0.06$  & $0.85 \pm 0.33$  & $0.56 \pm 0.48$  & $0.71 \pm 0.18$  & $0.59 \pm 0.17$     \\
\SpecFive    & $0.82 \pm 0.24$  & $0.81 \pm 0.24$  & $0.95 \pm 0.08$  & $0.83 \pm 0.25$  & $0.76 \pm 0.14$  & $0.53 \pm 0.09$     \\
\SpecSix    & $0.85 \pm 0.18$  & $0.88 \pm 0.18$  & $0.87 \pm 0.28$  & $0.85 \pm 0.28$  & $0.76 \pm 0.21$  & $0.68 \pm 0.17$     \\
\SpecOne    & $0.71 \pm 0.25$  & $0.89 \pm 0.07$  & $0.67 \pm 0.41$  & $0.97 \pm 0.01$  & $0.51 \pm 0.30$  & $0.56 \pm 0.15$     \\
All  & $0.86 \pm 0.20$  & $0.87 \pm 0.18$  & $0.90 \pm 0.24$  & $0.84 \pm 0.31$  & $0.74 \pm 0.19$  & $0.63 \pm 0.13$     \\
\noalign{\smallskip}\hline
\end{tabular}
\end{table*}
\subsection{Intraoperative Registration}\label{sec:supp_results_intraop_regi}
%
Full decompositions about each axis for pelvis pose errors are given Table~\ref{tab:results_decomp_pelvis_regi_errors} and highlight that nearly all of the pelvis translation error is in the projective depth direction.
%
%
\begin{table*}
\centering
\caption{Mean absolute values of each decomposed component of pelvis pose errors for which intraoperative registration was successful using method 3.
			 The axes are aligned with the projective coordinate frame with Z corresponding to depth.}
\label{tab:results_decomp_pelvis_regi_errors}       
\begin{tabular}{r r r r r r r}
\hline\noalign{\smallskip}
\multirow{2}{*}{Specimen} & \multicolumn{3}{c}{Rotation (\degree)} & \multicolumn{3}{c}{Translation (mm)} \\ 
& X  & Y & Z & X & Y & Z \\
\noalign{\smallskip}\hline\noalign{\smallskip}
\SpecTwo       & $0.1 \pm 0.1$ & $0.1 \pm 0.1$ & $< 0.1$  & $0.1 \pm 0.1$ & $0.1 \pm 0.1$ & $1.0 \pm 1.5$ \\
\SpecThree    & $0.1 \pm 0.1$ & $0.1 \pm 0.1$ & $< 0.1$  & $0.1 \pm 0.1$ & $0.1 \pm 0.1$ & $1.4 \pm 1.7$ \\
\SpecFour      & $0.1 \pm 0.1$ & $0.2 \pm 0.2$ & $0.1 \pm 0.1$  & $0.1 \pm 0.1$ & $0.1 \pm 0.1$ & $2.7 \pm 3.4$ \\
\SpecFive      & $0.1 \pm 0.1$ & $0.2 \pm 0.2$ & $0.1 \pm 0.1$  & $0.1 \pm 0.1$ & $0.1 \pm 0.1$ & $2.0 \pm 2.9$ \\
\SpecSix        & $0.1 \pm 0.1$ & $0.1 \pm 0.1$ & $< 0.1$  & $0.1 \pm 0.1$ & $< 0.1 $ & $0.9 \pm 1.2$ \\
\SpecOne      & $0.2 \pm 0.3$ & $0.2 \pm 0.1$ & $< 0.1$  & $0.1 \pm 0.1$ & $0.1 \pm 0.1$ & $3.0 \pm 3.2$ \\
All                  &	 $0.1 \pm 0.1$ & $0.1 \pm 0.1$ & $< 0.1$  & $0.1 \pm 0.1$ & $0.1 \pm 0.1$ & $1.4 \pm 2.0$ \\
\noalign{\smallskip}\hline
\end{tabular}
\end{table*}
\end{document}